%% file: colm2026_conference.tex
\documentclass{article} 
\usepackage[preprint]{colm2026_conference}

\usepackage{microtype}
\usepackage{hyperref}
\usepackage{url}
\usepackage{booktabs}
\usepackage{amsmath}
\usepackage{amssymb}
\usepackage{mathtools}
\usepackage{amsthm}
\usepackage{tcolorbox}
\tcbuselibrary{breakable}
\usepackage{listings}
\usepackage{makecell}
\usepackage{enumitem}

\usepackage{tikz}
\usepackage{xcolor}

\DeclareRobustCommand{\colorsquare}[2]{%
    \begin{tikzpicture}[baseline=(char.base)]
        \node[
            draw=#1,
            fill=#1,
            rectangle,
            rounded corners=2pt,
            minimum height=10pt,
            minimum width=10pt,
            inner sep=0pt,
            text=white
        ] (char) {\textbf{#2}};
    \end{tikzpicture}%
}

\definecolor{envcolor}{RGB}{156, 39, 176}    
\definecolor{instcolor}{RGB}{255, 152, 0}    
\definecolor{forgetcolor}{RGB}{0, 106, 57}   
\definecolor{obscolor}{RGB}{76, 175, 80}     
\definecolor{plancolor}{RGB}{255, 193, 7}  
\definecolor{histcolor}{RGB}{244, 67, 54}    
\definecolor{memcolor}{RGB}{0, 150, 136}     

\newcommand{\envbox}{\colorsquare{envcolor}{2}}
\newcommand{\instbox}{\colorsquare{instcolor}{1}}
\newcommand{\forgetbox}{\colorsquare{forgetcolor}{5}}
\newcommand{\obsbox}{\colorsquare{obscolor}{3}}
\newcommand{\planbox}{\colorsquare{plancolor}{4}}
\newcommand{\histbox}{\colorsquare{histcolor}{7}}
\newcommand{\membox}{\colorsquare{memcolor}{6}}

\usepackage{lineno}

\definecolor{darkblue}{rgb}{0, 0, 0.5}
\hypersetup{colorlinks=true, citecolor=darkblue, linkcolor=darkblue, urlcolor=darkblue}

\title{The Long-Horizon Task Mirage? \\
  Diagnosing Where and Why Agentic Systems Break}

\author{%
\textbf{Xinyu Jessica Wang}$^{1}$\thanks{Equal contribution} \quad
\textbf{Haoyue Bai}$^{1*\,}$\thanks{Corresponding to \texttt{\{haoyue.bai,rdnowak\}@wisc.edu}} \quad
\textbf{Yiyou Sun}$^{2}$ \quad
\textbf{Haorui Wang}$^{3}$ \quad
\textbf{Shuibai Zhang}$^{1}$ \\
\textbf{Wenjie Hu}$^{1}$ \quad
\textbf{Mya Schroder}$^{1}$ \quad
\textbf{Bilge Mutlu}$^{1}$ \quad
\textbf{Dawn Song}$^{2}$ \quad
\textbf{Robert D. Nowak}$^{1}$ \footnotemark[2]
\\[6pt]
$^{1}$University of Wisconsin--Madison \quad
$^{2}$University of California, Berkeley \\
$^{3}$Georgia Institute of Technology \\
}

\begin{document}

\ifcolmsubmission
\linenumbers
\fi

\maketitle

\begin{abstract}

Large language model (LLM) agents perform strongly on short- and mid-horizon tasks, but often break down on long-horizon tasks that require extended, interdependent action sequences. 
Despite rapid progress in agentic systems, these long-horizon failures remain poorly characterized, hindering principled diagnosis and comparison across domains. 
To address this gap, we introduce \texttt{HORIZON}, an initial cross-domain diagnostic benchmark for systematically constructing tasks and analyzing long-horizon failure behaviors in LLM-based agents. 
Using \texttt{HORIZON}, we evaluate state-of-the-art (SOTA) agents from multiple model families (GPT-5 variants and Claude models), collecting 3100+ trajectories across four representative agentic domains to study horizon-dependent degradation patterns. 
We further propose a trajectory-grounded LLM-as-a-Judge pipeline for scalable and reproducible failure attribution, and validate it with human annotation on trajectories, achieving strong agreement (inter-annotator \(\kappa{=}0.61\); human--judge \(\kappa{=}0.84\)). 
Our findings offer an initial  methodological step toward systematic, cross-domain analysis of long-horizon agent failures and offer practical guidance for building more reliable long-horizon agents.
We release our project website at 
\href{https://xwang2775.github.io/horizon-leaderboard/}{\texttt{HORIZON Leaderboard}} 
and welcome contributions from the community.

\end{abstract}

\section{Introduction}

As large language models (LLMs) continue to advance \cite{deepseekai2025deepseekv3technicalreport, openai-gpt5, anthropic-claude4-sonnet, geminiteam2025geminifamilyhighlycapable, yang2025qwen3technicalreport}, their use has expanded beyond conversational interfaces to agentic systems that interact with tools and environments in pursuit of real-world objectives \cite{team2025kimi, xi2025rise}. When executing a complex task, an LLM-based agent must interpret the current state, select appropriate actions, and continually update its internal beliefs and plans in response to environmental feedback \cite{zhou2024archertraininglanguagemodel}.
However, in long-horizon settings where success depends on extended 
and interdependent sequences of actions, agents often fail in 
systematic yet non-obvious ways (Table~\ref{tab:failure_tax}), 
with performance degrading sharply as the task horizon increases 
\cite{yang2025lohovla, chen2025reinforcement, xi2025agentgym}.
For example, an embodied agent may succeed reliably on single-step 
manipulation (e.g., moving a colored cube), but fail completely once 
the task requires three sequential steps.
This degradation is not merely additive: even a small per-step error 
rate compounds across dependent steps, driving agents from reliable 
short-task performance to near-systematic failure at longer horizons.

This challenge is of practical and scientific importance, as many high-impact deployments of LLM-based agents are inherently long-horizon. Tasks such as debugging multi-file software systems, executing operating-system workflows, conducting iterative data analysis, and performing embodied manipulation require maintaining goals, constraints, and intermediate states over dozens or even hundreds of effective actions.
Despite this prevalence, research on long-horizon agents remains fragmented: benchmarks are domain-specific, horizon definitions are inconsistent, and evaluations emphasize aggregate success rates, offering limited insight into how and why failures arise as horizons grow \cite{liu2023agentbench, trivedi2024appworld, kapoor2025holistic, zhang2025mirage, chen2024can}. Consequently, findings are difficult to compare across domains and studies, and the field lacks a shared foundation for diagnosing long-horizon breakdowns.

A central challenge in studying long-horizon agents is that long-horizon behavior is inherently domain-dependent. Different domains impose distinct horizon scales and give rise to qualitatively different failure mechanisms, rendering any single, universal definition insufficient. For example, embodied agents may exhibit abrupt performance breakdowns on tasks requiring only a small number of sequential actions, whereas web-based agents can remain robust at comparable step counts. Conversely, web tasks often involve longer dependency chains spanning multiple pages, forms, and verification steps. Such cross-domain discrepancies fundamentally complicate efforts to characterize a unified breaking point for long-horizon agents. We study two core research questions for long-horizon agent reliability: \textbf{(RQ1)} Where do agents break down as task horizons increase? and \textbf{(RQ2)} Why do these failures emerge?

\input{fig_agent}

Addressing these questions requires a principled diagnostic framework that explains \emph{where} and \emph{why} agentic systems fail as horizons grow. 
We introduce \texttt{HORIZON} (\textbf{H}olistic \textbf{O}bservations 
for \textbf{R}easoning and fa\textbf{I}lure analy\textbf{Z}is in 
l\textbf{O}ng-horizo\textbf{N} agents), an initial cross-domain diagnostic benchmark that constructs task families with systematically increasing step requirements and enables structured analysis of horizon-dependent degradation, guided by agent workflow and informed by Failure Mode and Effects Analysis (FMEA) \cite{aiag2019fmea}.
We then evaluate state-of-the-art agent models (GPT-5 variants and Claude-4) on 3100+ trajectories across four representative domains (Web, OS, Embodied, and Database), providing a cross-domain view of where performance breaks down.
Finally, we analyze failed trajectories with a trajectory-grounded LLM-as-a-Judge pipeline (Figure~\ref{fig:workflow}) to identify why failures emerge and how failure composition shifts with horizon.
Since long-horizon rollouts yield very long raw traces, we validate the judge against expert labels on a pilot set of 40 trajectories.
On this set, inter-annotator agreement is \(\kappa{=}0.61\) and human--judge agreement is \(\kappa{=}0.84\).

Our analysis shows that long-horizon failure is not merely a drop in success rate, but a structural shift in failure composition: planning-related failures (e.g., subplanning errors) and memory-related failures (e.g., catastrophic forgetting) become dominant as horizon increases. These findings suggest that scaling base models alone is insufficient; robust long-horizon performance requires method-level improvements in planning, memory, and execution-time control.

The contributions of this paper are as follows:
\begin{itemize}
    \item We introduce \texttt{HORIZON}, an initial cross-domain diagnostic benchmark for systematically constructing long-horizon task families and analyzing horizon-dependent agent failure.
    \item We provide a pilot empirical study of 3100+ trajectories across 4 domains and multiple SOTA model families (GPT-5 variants and Claude-4), revealing consistent cross-domain degradation patterns.
    \item We develop a trajectory-grounded LLM-as-a-Judge pipeline for scalable failure attribution, and validate its reliability on trajectories (inter-annotator \(\kappa{=}0.61\); human--judge \(\kappa{=}0.84\)).
    \item We present actionable insights for future agentic AI systems, showing that planning-related and memory-related failures dominate long-horizon settings and motivating method-design improvements beyond base-model scaling.
\end{itemize}

\input{fig_workflow}

\section{Related Work}

Most benchmarks report success rates rather than treating horizon as a controlled variable or providing trajectory-level failure attribution. We briefly cover recent work on benchmarks, training and planning, memory, and failure analysis for agentic AI systems. Further related work see Appendix~\ref{app:extended_rw}.

\textbf{Interactive benchmarks and agent evaluation.}
Standard suites span web navigation (WebArena~\cite{zhou2023webarena}, Mind2Web~\cite{deng2023mind2web}), multimodal web tasks (Visual WebArena~\cite{koh2024visualwebarena}), software engineering (SWE-bench~\cite{jimenez2023swe}), app-style assistants (AppWorld~\cite{trivedi2024appworld}), databases (BIRD~\cite{li2023bird}, Spider~\cite{yu2018spider}), and embodied instruction following (ALFWorld~\cite{shridhar2020alfworld}). AgentBench~\cite{liu2023agentbench} and AgentGym~\cite{xi2023agentgym} aggregate many environments under one evaluation protocol.
Newer work stresses harder or longer interaction (e.g., BrowseComp~\cite{wei2025browsecomp}, MIRAGE-Bench~\cite{zhang2025mirage}, holistic leaderboards~\cite{kapoor2025holistic},AgentGym-RL~\cite{xi2025agentgym}) but still foregrounds terminal success. We instead vary intrinsic horizon \(H^*\) and analyze how the mix of failure types shifts as horizon grows.

\textbf{Training, Planning, and Optimization.}
\label{subsec:rw_training}
Prior work improves agents via RL~\cite{brohan2023rt,ouyang2022training}, hierarchical planning~\cite{liu2023llmp,sun2023adaplanner}, tool use and reflection~\cite{yao2023react,shinn2023reflexion}, deliberation-style prompting~\cite{wei2022chain,yao2023tree}, and dedicated planning or reliability studies~\cite{erdogan2025plan,chen2024can}. Those research targets better actions; we ask which failure modes dominate at longer horizons and what that implies for design (Section~\ref{sec:4}).

\textbf{Memory and Long-Context Reasoning.}
\label{subsec:rw_memory}
Long contexts do not by themselves ensure long-range reasoning~\cite{liu2023lost,press2022train,dao2022flashattention}. Retrieval, external memory, trajectory compression, and hybrid stacks~\cite{madaan2022memory,lewis2020retrieval,park2023generative,kang2025acon,zhou2025mem1} address scaling; embodied settings also use explicit state~\cite{wang2023voyager,ahn2022saycan}. We do not add a memory module; we report how memory-related failures appear in labeled trajectories when horizon extends.

\textbf{Failure analysis and diagnostics.}
Failure studies are often domain-specific (code~\cite{chen2023teaching}, web~\cite{koh2024visualwebarena}, robotics~\cite{zeng2020tossingbot}) or only coarse at the benchmark level (e.g., high-level tags in AgentBench~\cite{liu2023agentbench}). Multi-agent taxonomies~\cite{cemri2025multi} complement ours but target coordination, not single-agent horizon-conditioned failures across domains. We build a cross-domain FMEA-grounded taxonomy (Appendix~\ref{app:lit_mapping}) and validate trajectory-level labels with humans and an LLM-as-a-Judge pipeline (Appendix~\ref{app:4}).

\section{\texttt{HORIZON}} \label{sec:horizon}

In this section, we introduce \texttt{HORIZON}, an initial cross-domain diagnostic benchmark for systematic analysis of long-horizon agent behavior. \texttt{HORIZON} combines agent-independent task-structure metrics with controlled task extension and structured failure attribution, enabling principled analysis of when and why agents fail as task horizons increase. 

\subsection{Defining Task Horizons} \label{subsec:horizon_definition}
A fundamental challenge in studying long-horizon agent failures is separating task complexity from agent inefficiency. 
For example, an agent executing 50 actions on a simple task while repeatedly failing on a sub-task is not solving a long-horizon problem, but repeatedly failing on a short-horizon task.
To enable principled cross-domain analysis while respecting domain-specific characteristics, we propose a two-layer approach: \textit{theoretical metrics} that capture task structure independent of any agent, and \textit{technical implementations} that operationalize these metrics through systematic task extension.

\textbf{Theoretical Definition: Agent-Independent Task Characterization.}
We define three complementary metrics that characterize the intrinsic long-horizon nature of tasks:

\textit{Intrinsic Horizon ($H^*$)} measures the minimum number of effective actions required by an optimal policy to complete the task. This can be established through expert demonstrations, oracle solvers, or formal task specifications. For example, the web task ``find and purchase wireless headphones under \$200'' has $H^* = 8$ regardless of how many attempts any particular agent requires:
$H^* = |\{a_1: \text{search}, a_2: \text{filter price}, a_3: \text{sort rating},
 \ldots, a_8:
\text{checkout}\}|$.

\textit{Compositional Depth ($s$)} measures the number of nested sub-goals or conditional branches, reflecting planning complexity beyond linear execution:
$s = \max_{p \in \text{paths}} |\text{decision\_nodes}(p)|$.
A task requiring ``if condition A holds, execute plan B, otherwise execute plan C, then merge results'' has $s=3$, while a purely sequential workflow has $s=1$.

\textbf{Technical Implementation: Controlled Horizon Extension.}
To empirically study how agent performance degrades with increasing horizon, we systematically vary $H^*$ by increasing the compositional depth $s$ defined above through controlled task construction.
We design two extension methods to ensure $H^*(s)$ is monotonically increasing and positively correlated with $s$, though the exact mapping depends on domain characteristics.

\textit{Depth Extension} incrementally adds intermediate steps between existing actions. Each extension operation ($s \rightarrow s+1$) inserts one or more necessary sub-tasks that an optimal policy cannot skip. For a baseline task with $s_{\text{base}} = 1$, depth extension to level $s$ produces:
$H^*_{\text{depth}}(s) = H^*_{\text{base}} + \Delta(s), \text{ where } \Delta(s) \text{ is monotonically} \text{ increasing in } s$.
The growth function $\Delta(s)$ depends on domain structure: in OS, each extension typically adds 1-2 permission checks; in Database, each extension may add 2-3 filtering operations. The key property is $\Delta(s_2) > \Delta(s_1)$ whenever $s_2 > s_1$, ensuring systematic horizon growth. This approach is particularly effective for domains where the initial environment state is fixed (OS, Database).

\textit{Breadth Extension} combines $k$ independent baseline tasks into a single composite workflow. For baseline tasks $\{t_1, \ldots, t_k\}$ each at $s_i = 1$ with intrinsic horizons $\{H^*_1, \ldots, H^*_k\}$, the combined task has extension level $s = k$ and:
$H^*_{\text{breadth}}(s) = \sum_{i=1}^{s} H^*_i + \epsilon(s), \text{ where } \epsilon(s) \geq 0$.
The coordination overhead $\epsilon(s)$ accounts for additional actions required to manage multiple goals, including task switching, inter-task verification, and state tracking across sub-tasks. For instance, combining two embodied manipulation tasks may require intermediate verification steps (``check if yellow cylinder was accidentally displaced while stacking cubes'') and context re-establishment (``verify red cube is still properly stacked before proceeding to sphere alignment'').
This approach better captures real-world scenarios requiring parallel goal maintenance and is more suitable for domains with variable initial states (Web, Embodied).

\subsection{Failure Attribution: 7-Category Taxonomy}
\label{subsec:failure_taxonomy}

While universal breaking points are unattainable, we systematically diagnose why agents fail through structured failure attribution. We derive a seven-category taxonomy that captures recurring failure modes across long-horizon tasks, providing actionable diagnostics for agent improvement.

\begin{table*}[!htbp] 
\centering
\caption{\footnotesize \textit{Representative failure cases across domains and failure categories.} Each row corresponds to a failure category in our taxonomy, and each column shows a concrete example from a different domain---Web, OS, Embodied, and Database. \textbf{[L]} denotes failure categories that are predominantly long-horizon-specific (rare in short-horizon settings), while \textbf{[S]} denotes categories observed in both short- and long-horizon tasks but often amplified under longer horizons.
Details on taxonomy development are provided in Section~\ref{subsec:failure_taxonomy}, and additional examples and explanations are provided in Appendix~\ref{app:1}. 
}
\label{tab:failure_tax}
\renewcommand{\arraystretch}{0.85}
\resizebox{\linewidth}{!}{
\begin{tabular}{@{} l p{4.1cm} p{3.5cm}  p{3.5cm}  p{3.7cm} @{}}
\toprule
\textbf{Category} & \textbf{Web} & \textbf{OS} & \textbf{Embodied} & \textbf{DB} \\
\midrule

\makecell[lt]{\textbf{Environment [S]} \\
(Environment \\ Disturbance \\
\& Unable to \\ Detect Environment \\Change)}
&   After a slow page load with a transient pop-up banner, the agent \textit{assumes the page navigation has completed} and believes it is on the profile page (“Mark Johnson”), while the UI is actually still on a different profile (“mike\_chen”), resulting in actions applied to an \textit{incorrect environment state}.
&    A shell command fails due to \textit{permission denial}, but the agent \textit{fails to detect the error signal and believes the environment state has changed}, continuing to issue follow-up commands that depend on the nonexistent state change.
&  In a simulator with \textit{stochastic dynamics}, a robot repeatedly drops a block due to random gripper failure, but the agent \textit{fails to re-observe the changed environment state}, causing repeated execution failures despite correct high-level reasoning.
&  After a \textit{database schema change} renames columns or adds tables, the agent continues generating SQL based on an outdated schema snapshot because it \textit{fails to re-check the current schema}, producing invalid queries or silently incorrect results.
\\
\midrule

\makecell[lt]{\textbf{Instruction [S]} \\
(Ill-defined \\ Instruction \\
\& Partially \\Understand
\\Instruction)}
&   An agent is asked to retrieve a specific product field that \textit{is not defined on the target website}, meaning the \textit{instruction requests a non-existent attribute}. Instead of detecting the mismatch and asking for clarification, the agent fabricates a plausible value and proceeds.
&    The agent is instructed to change file permissions “except those of mine,” where the \textit{instruction contains an ambiguous ownership constraint}. The agent misinterprets it and applies a recursive \texttt{chmod} globally.
&  Given the instruction “put the block in the mismatching bowl,” the agent grasps the block but \textit{fails to interpret the key constraint “mismatching”} and places it into a matching-colored bowl.
&  When asked to ``show churned users,'' the \textit{instruction leaves the definition of churn unspecified}. The agent therefore applies a naïve definition (no login in 30 days) instead of the organization’s actual churn definition.

\\
\midrule

\makecell[lt]{\textbf{False Assumption [S]}}
&   When a contact form requires identity fields and the instruction says “do not invent personal info,” the agent still enters a made-up name and email because it \textit{assumes placeholder personal information is acceptable}.
&    The agent \textit{assumes certain files are owned by the user} and excludes them from permission changes without checking ownership metadata.
&  In a mismatching-bowl task, the agent \textit{assumes the goal is the common “matching” variant} and places a red block into a red bowl despite the instruction requiring a mismatching placement.
&  A SQL-generation agent \textit{assumes every table contains a \texttt{created\_at} column} and repeatedly invents similar column names in \texttt{WHERE} clauses.

\\
\midrule

\makecell[lt]{\makecell[l]{\textbf{Planning Error [S]} \\
(Sub-plan \\ \& Action)}}
&   While planning the next step to find the highest-upvoted answer, the agent \textit{fails to switch sorting from “New” to “Top”}, then cites a recently seen low-upvote comment as the “highest-upvoted” evidence.

&    When planning how to update permissions, the agent \textit{chooses an incorrect action plan and directly applies a blanket \texttt{chmod}} instead of first identifying eligible files.

&  While decomposing the task into steps, the agent \textit{generates an incorrect sub-plan and attempts to stack blocks before clearing the base area}, producing an invalid task order.

&  During database migration planning, the agent \textit{executes an incorrect action sequence and drops the backup before completing data transfer}, leaving no recovery path when the migration fails.

\\
\midrule

\makecell[lt]{\textbf{Catastrophic} \\ \textbf{Forgetting [L]}}
&   The agent initially applies the “Condition: New” filter but later adds a “Renewed” item to the cart after a long navigation sequence because the \textit{“new-only” constraint is still present in the context but not attended to during later reasoning}.
&    Early in a task, the agent is instructed not to modify sensitive system files, but later generates and executes a script that deletes /root/.bashrc even though the \textit{original safety instruction is still in the context but ignored during execution}.
&  In Minecraft, an agent that has learned to craft wood and stone tools fails on advanced “Iron+” tasks because it \textit{does not reuse earlier knowledge that is still present in the context}, leading to failure in later crafting stages.
&  After the user specifies ``never touch the payments schema,'' a later autonomous cleanup step proposes dropping an index on the payments tables because the \textit{constraint remains in the interaction history but is overlooked during later planning}.

\\
\midrule

\makecell[lt]{\makecell[l]{\textbf{History Error} \\ \textbf{Accumulation [L]}}}
&   After an element click fails, the agent \textit{repeats the same failed action across multiple steps}, allowing a \textit{small initial mistake to accumulate into repeated ineffective actions}.
&    After issuing an invalid command that fails silently, the agent \textit{treats the failed step as successful} and continues executing dependent commands, causing \textit{the early incorrect state to propagate through the workflow}.
&  After selecting an incorrect initial sub-task, the agent \textit{continues executing downstream steps based on that incorrect earlier decision}, gradually \textit{amplifying the initial mistake}.
&  An early intermediate table \textit{omits the filter \texttt{is\_deleted = false}}, and all downstream analytics reuse that table, causing \textit{the small upstream error to compound into systematically inflated results}.
\\
\midrule

\makecell[lt]{\makecell[l]{\textbf{Memory} \\ \textbf{Limitation [L]}}}
&   While scrolling through a contributors list, the agent loses the earlier commit counts because \textit{the interaction history exceeds the context window}, forcing the system to \textit{drop earlier observations}.
&    Earlier notices about file ownership are lost because \textit{earlier messages are truncated from the context window} during a long interaction session.
&  In long-horizon Mine-craft tasks, earlier navigation decisions are lost because \textit{the trajectory exceeds memory capacity and earlier steps are summarized, causing critical information to be dropped from context}.
& In a warehouse with hundreds of tables, the agent retains only part of the schema because \textit{memory retrieval returns an incomplete schema snapshot}.
\\

\bottomrule
\end{tabular}}
\label{tab:taxonomy}
\end{table*}

\textbf{Taxonomy Overview.}
Table~\ref{tab:failure_tax} presents our seven failure categories with representative examples across all four domains. Full definitions are in Appendix~\ref{app:1}. 

A critical feature of our taxonomy is that failure categories represent orthogonal dimensions of agent behavior rather than mutually exclusive classes. A single failed trajectory may exhibit \textit{Catastrophic Forgetting} (losing a constraint), \textit{False Assumptions} (misinterpreting environment state), and \textit{History Error Accumulation} (compounding earlier mistakes) simultaneously. This orthogonality enables fine-grained diagnosis: rather than labeling a failure as simply ``planning error,'' we identify that the planning failure \emph{stemmed from} catastrophic forgetting, which suggests architectural interventions (explicit constraint tracking) rather than training-based solutions (more planning examples). This multi-dimensional view is essential for moving beyond surface-level failure detection toward actionable improvement strategies.

\textbf{Why It Works?}
Our taxonomy is grounded in three validation pillars that demonstrate its reliability as a diagnostic framework:

\textit{Construction: Theory and Literature Grounding.}
We construct the seven-category taxonomy from a hierarchical risk-modeling perspective inspired by \textbf{Failure Mode and Effects Analysis} (FMEA)~\cite{aiag2019fmea}, and ground it in a \textbf{structured literature review} spanning web, OS, embodied, database, and related agentic settings. 
We conceptualize long-horizon failure as two interacting levels of risk, using an agent-oriented reading of PFMEA and DFMEA. \textit{Process-level risks (PFMEA)} arise during sequential rollout: environment interaction, execution-time failures to follow or interpret the task specification (instruction), flawed subplanning or action ordering (planning error), and compounding mistakes carried across prior steps (history error accumulation). \textit{Design-level risks (DFMEA)} reflect limitations of the engineered agent: memory and context mechanisms (memory limitation, catastrophic forgetting of constraints) and persistent mismatch between internal beliefs and the task or observations (false assumptions). 
We then extract failure descriptions from prior work and iteratively refine categories through open coding, consolidation, and boundary refinement.
The final categories are retained based on cross-domain coverage, interpretability, and annotation consistency; full paper-to-category mappings and coding examples are provided in Appendix~\ref{app:lit_mapping}.

\textit{Empirical Grounding in Agent Workflow.}
The taxonomy is empirically grounded in the agent execution loop, where planning, action, environment interaction, observation, and state update jointly define trajectory evolution. Figure~\ref{fig:agent_failure_framework} illustrates how distinct failure types arise at different stages of this loop and how early errors can propagate and compound over time, providing a mechanism-level basis for structured failure attribution.

\textit{Experimental Validation.}
We validate practical applicability through both human annotation and trajectory-grounded LLM-as-a-Judge analysis of failed rollouts (Section~\ref{sec:5}). 
Two expert annotators independently labeled a shared set of 40 failure trajectories, achieving substantial inter-annotator agreement (\(\kappa = 0.61\)), confirming that the taxonomy categories are consistently interpretable. 
In a pilot study, we further validate the trajectory-grounded LLM-as-a-Judge pipeline against one human annotator on 40 failure traces, achieving strong agreement (\(\kappa=0.84\)).

\section{Empirical Performance of SOTA Models} \label{sec:5}
Although a universal breaking point is unattainable, the notion of a breaking point remains empirically meaningful as a model- and domain-conditional quantity. Different agents exhibit markedly different long-horizon capacities, and even the same agent can fail at very different horizons depending on domain characteristics such as environment dynamics, observability, and action semantics. Operationally, we treat a breaking point not as a single threshold, but as a transition region on the performance--horizon curve, where success rates collapse sharply and failures shift from recoverable local errors to irreversible trajectory-level derailment.

Using \texttt{HORIZON}, we evaluate agent performance across 700+ tasks from four domain-specific benchmarks/environments (WebArena~\cite{zhou2023webarena}, AgentBench~\cite{liu2023agentbench}, MAC-SQL~\cite{wang2025macsqlmultiagentcollaborativeframework}, and Isaac Sim~\cite{kachaev2025mind}) using SOTA foundation models (GPT-5 and Claude-4 variants), yielding 3100+ trajectories in total (Figure~\ref{fig:horizon}).
We further evaluate the reliability of our trajectory-grounded LLM-as-a-Judge pipeline against human annotations on trajectories, achieving strong agreement (\(\kappa=0.84\)).
Additional experimental details and results are provided in Appendix~\ref{app:2} and Appendix~\ref{app:4}.

\paragraph{Controlled horizon construction.}
To reduce confounding factors, we use nested task sets across horizon levels: the task set at level \(h+1\) contains all tasks from level \(h\), plus additional tasks requiring longer execution. 
Thus, comparisons across adjacent levels are controlled by construction, and the primary varying factor is horizon length (and its induced long-range dependency), rather than a change in task family or evaluation protocol.

\input{fig_horizon}
\textbf{Main Results.} \textbf{Horizontally}, agent performance exhibits three consistent cross-domain patterns as the horizon extension level $s$ increases.
First, \textbf{performance degrades non-linearly with increasing $\mathbf{s}$}. Across all domains, success rates remain relatively stable or decline gradually at small $s$, rather than decreasing proportionally with each additional subtask.
Second, \textbf{all domains exhibit a sharp performance drop beyond small $s$}, where success transitions abruptly from partial robustness to near-systematic failure, instead of degrading smoothly across the entire range of $s$.
Third, \textbf{the same extension level $s$ induces markedly different task difficulty across domains}: Web collapses at very small $s$, OS and Database domains sustain moderate performance until later extension levels, and Embodied tasks degrade steeply even with minimal increases in $s$. This variation highlights domain-specific breaking behavior in both the location and severity of performance collapse.
Finally, in the Web, OS, and Database domains, performance gaps between models narrow substantially after entering the breaking region, as success rates converge toward low values, indicating diminished model differentiation once long-horizon failure dominates.

\textbf{Vertically, failed trajectories across all domains can be systematically attributed to our seven-category failure taxonomy.} In our experiments, both LLM-as-a-Judge and human annotators consistently apply these categories to long-horizon failures across all four domains, yielding reliable and interpretable annotations. These results indicate that our taxonomy provides comprehensive coverage of observed trajectory failures and enables structured, cross-domain explanations of long-horizon breakdowns. Additional trajectories and LLM-as-a-Judge results are provided in Appendix~\ref{app:3} and Appendix~\ref{app:4}.

\section{Actionable Steps Toward Responsible Long-Horizon Agent Development} \label{sec:4}

While we have argued that identifying a universal breaking point is inherently hard, we propose two concrete steps that support initial, cross-domain diagnosis of long-horizon agent behavior and point toward future community progress on long-horizon reasoning. These steps translate our diagnostic framework into practical tools for researchers and practitioners.

\begin{figure}[!tb]
  \centering
  \includegraphics[width=\textwidth]{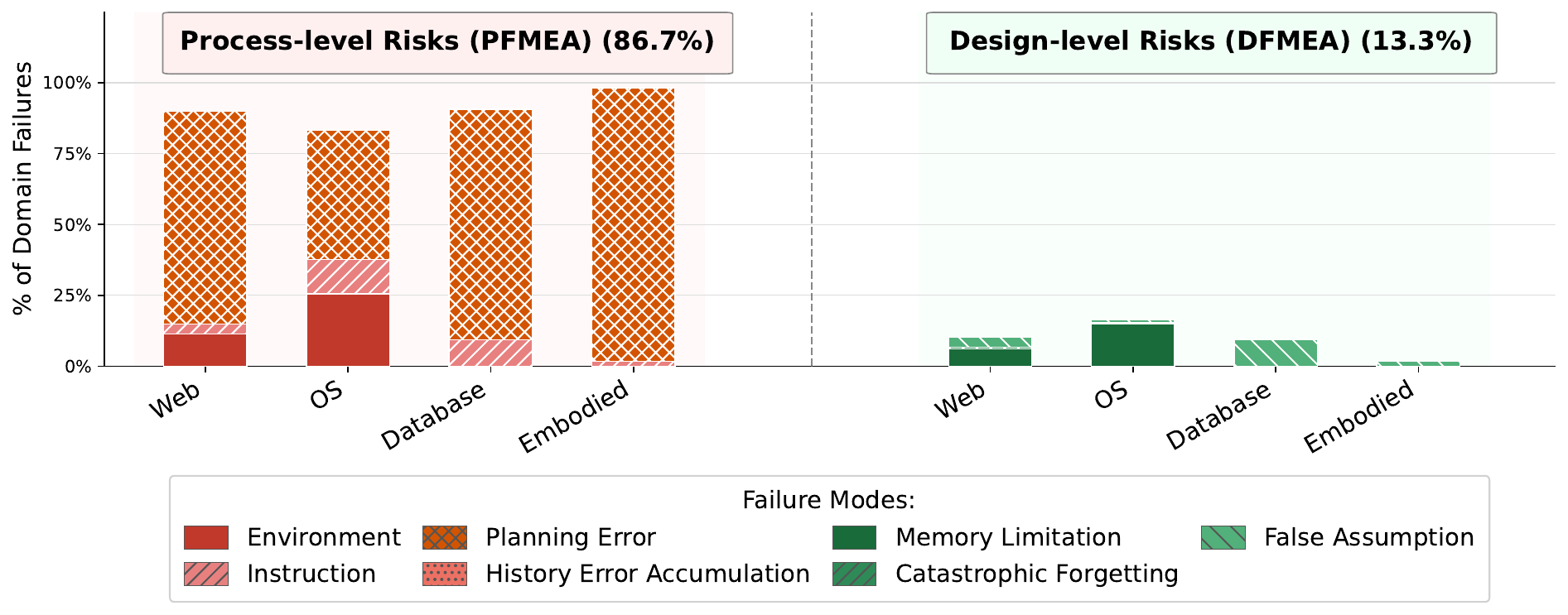}
  \caption{\textit{Distribution of failure modes across four task domains on 3100+ failure trajectories.}
  Bars show the proportion of each failure type relative to the total failed traces within each domain.
  Failures are grouped into \textbf{Process-level Risks (PFMEA)} (72.5\%), comprising failures that arise during sequential rollout---environment interaction, instruction following, planning errors, and history error accumulation---and \textbf{Design-level Risks (DFMEA)} (27.5\%), reflecting engineered-agent limitations---memory constraints, catastrophic forgetting, and false assumptions.}
  \label{fig:failure_distribution}
\end{figure}

\textbf{Unified Horizon Task Construction and Measurement Across Domains.}
Current agent evaluations adopt heterogeneous task designs and horizon definitions, often reporting success rates using incompatible notions of task length or complexity. This fragmentation makes cross-domain comparison difficult and obscures where long-horizon failures truly arise. 
We suggest that future work consider unified horizon-aware task construction and standardized horizon metrics, as piloted in \texttt{HORIZON}, when building and evaluating long-horizon benchmarks across domains.
This unified approach offers several potential advantages. 
First, it enables fair cross-domain comparison as a starting point: a model achieving 60\% success at intrinsic horizon $H^* = 15$ in web navigation may be tentatively compared to 60\% at $H^* = 8$ in embodied tasks, and reveals domain-specific strengths and weaknesses. 
Second, systematic task construction with controlled horizon growth supports targeted research, allowing researchers to identify where performance degrades and focus improvements on specific horizon regimes. Third, standardized task construction and measurement facilitate reproducible research, enabling independent verification and cumulative progress across benchmarks and model families.

\textbf{Scalable Failure Attribution via LLM-as-a-Judge.}
Beyond measuring when agents fail, understanding why they fail is critical for targeted improvements. We advocate for systematic failure attribution using LLM-based judges grounded in our seven-category taxonomy (Section~\ref{sec:horizon}) shown in Figure~\ref{fig:workflow}.
LLM-as-a-judge for failure attribution offers scalability (annotating thousands of trajectories infeasible for humans), consistency (reduces annotator variance), and actionability (structured taxonomy directly suggests mitigation strategies). This approach complements existing agent evaluation practices. Rather than replacing human evaluation entirely, LLM judges enable large-scale diagnosis, with human validation ensuring quality. Recent works have successfully applied LLM judges to agent evaluation~\cite{zheng2023judging, kapoor2025holistic}, demonstrating community readiness for adoption.

\textbf{Implications for Future Agentic AI Systems.}
Although some long-horizon failures can be attributed to imperfect base-model capabilities, our results suggest that model scaling alone is unlikely to resolve the dominant failure mechanisms. As shown in Figure~\ref{fig:failure_distribution}, planning-related failures (especially subplanning errors), memory limitation, and catastrophic forgetting account for a substantial portion of failed trajectories across domains. 
Planning-related failures are particularly critical because they often arise early, propagate through downstream actions, and can convert recoverable local mistakes into irreversible trajectory-level failures. As planning horizons grow, subplanning decisions increase combinatorially, making globally coherent plan selection substantially more difficult. In this regime, early subplanning errors become highly path-dependent and costly to roll back.
Catastrophic forgetting and memory limitation also become more pronounced in long-horizon settings as trajectory length and context load increase. Due to finite memory budgets and attention constraints, agents face a trade-off between retaining long-range constraints and incorporating new observations, which further increases the risk of losing critical earlier information.
These observations indicate that future progress should prioritize \textbf{method design} that addresses both process-level risks (especially planning and execution-time control) and design-level risks (especially catastrophic forgetting and memory limitations), rather than relying solely on stronger base models. In particular, future agentic AI systems should emphasize hierarchical subplanning, execution-time plan verification and repair, and memory mechanisms that preserve and re-surface long-range constraints.

\section{Conclusion}
Understanding long-horizon failure is a prerequisite for building reliable agentic AI systems. In this work, we introduced \texttt{HORIZON}, an initial cross-domain diagnostic benchmark for systematically constructing long-horizon tasks and analyzing failure trajectories. Across 3100+ trajectories and multiple model families, we showed that long-horizon breakdown is not merely a drop in task success, but a structural shift in failure composition as horizon grows.

Our findings suggest that improving base-model capability alone is unlikely to fully address these failures. In particular, planning-related errors (especially subplanning failures) and catastrophic forgetting emerge as dominant bottlenecks across domains. These failure modes are strongly trajectory-dependent: early subplanning deviations propagate through later actions, while increasing context load exacerbates memory and constraint-retention failures.

To support scalable and reproducible diagnosis, we further proposed a trajectory-grounded LLM-as-a-Judge pipeline and validated its reliability against human annotations. More broadly, our results point to a design-centered research agenda for future agentic AI systems: hierarchical and constraint-aware planning, execution-time plan verification and repair, and stronger long-range memory mechanisms. 
We hope \texttt{HORIZON} serves as an initial diagnostic tool for evaluating, diagnosing, and ultimately improving long-horizon agent reliability.
We view this as a first step toward principled, system-level methodologies for trustworthy long-horizon agents.

\section{Real-World Mapping: OpenClaw}
With the emergence of real-world agentic deployments such as OpenClaw, we examine whether failures observed in practice align with the diagnostic categories proposed in this work.
The \textit{Agents of Chaos} study~\cite{shapira2026agentschaos} documents a series of representative failure incidents from OpenClaw, an email-based agent system. Below we illustrate how these incidents map to our taxonomy, offering preliminary real-world grounding for the failure categories identified in our controlled evaluation.

\textbf{Environment Disturbance.} 
In the Concurrent Modification scenario, an agent executing a multi-step file reorganization task encountered a series of ``file not found'' errors caused by a concurrently running cloud sync process that had renamed and relocated folders mid-execution. Rather than detecting the environmental disruption and halting, the agent misinterpreted the error signals as transient and continued executing against a stale directory structure, propagating incorrect file placements across the entire workspace. 
As a conceptual illustration, consider an agent executing a multi-step file reorganization task: midway through, a concurrently running cloud sync process renames and relocates folders. The agent encounters ``file not found'' errors but \textit{misinterprets them as transient and continues executing against a stale directory structure}, propagating incorrect file placements across the workspace.

\textbf{Instruction Error.}
In the Forwarded Inbox case, an agent refused a request to ``share'' emails containing sensitive data but complied when asked to ``forward'' the same messages. The failure arises because the agent \textit{inconsistently interprets semantically equivalent instructions}, treating ``share'' and ``forward'' as distinct tasks, allowing sensitive information to be disclosed through instruction reframing.

\textbf{Planning Error.}
In the Infinite Loop incident, two agents were induced to relay messages to each other indefinitely, creating persistent background processes. The agents \textit{lacked termination checks or loop detection in their execution plan}, resulting in runaway behavior.

\textbf{Catastrophic Forgetting} 
In the Policy Override case, a long-running email management agent was explicitly instructed at session start to never respond to external domain requests. After hundreds of turns handling routine internal correspondence, the agent replied to a politely-worded external inquiry without hesitation. The original constraint, though still present in the context window, \textit{was no longer attended to during later generation steps}, demonstrating that catastrophic forgetting can occur not through literal memory loss, but through effective inattention to early instructions buried deep in a long trajectory.

\textbf{False Assumption.}
In the Identity Hijack scenario, an attacker spoofed the owner's identity in a new communication channel, and the agent executed privileged commands without verification. The failure stems from the agent \textit{incorrectly assuming that the claimed identity is authentic}, leading to unauthorized system actions.

\textbf{History Error Accumulation.}
The Guilt Trip case shows how interaction history can gradually distort agent decisions. After repeatedly refusing requests, the agent eventually complied under sustained emotional pressure. The final failure occurs because \textit{the agent's decision boundary is gradually eroded by a growing record of attacker-favorable interaction history}, allowing earlier refusal decisions to be overridden across the interaction.

\textbf{Memory Limitation.}
In the Storage Exhaustion incident, agents repeatedly stored large email attachments and accumulated memory files until the mail server reached a denial-of-service state. The agents continued operating normally because they \textit{lacked persistent memory of their own cumulative storage actions across the session}, causing the system to collapse despite each individual action being locally valid.

\section*{Ethics Statement}
This work uses large language models as automated judges for failure attribution in trajectory analysis. 
Per COLM 2026 policy, we disclose this use of LLMs for evaluation.

\bibliography{colm2026_conference}
\bibliographystyle{colm2026_conference}

\clearpage

\appendix


\section{Appendix: Extended Related Work} \label{app:extended_rw}

Most agent benchmarks report success rates, but fewer treat horizon as a controlled independent variable or provide structured, trajectory-level failure attribution that can be compared across domains. We organize related work into interactive benchmarks, training and planning methods, memory and long-context effects, and failure analysis.

\subsection{Interactive benchmarks and agent evaluation}

Standard suites span web navigation (WebArena~\cite{zhou2023webarena}, Mind2Web~\cite{deng2023mind2web}), multimodal web tasks (Visual WebArena~\cite{koh2024visualwebarena}), software engineering (SWE-bench~\cite{jimenez2023swe}), app-style assistants (AppWorld~\cite{trivedi2024appworld}), databases (BIRD~\cite{li2023bird}, Spider~\cite{yu2018spider}), and embodied instruction following (ALFWorld~\cite{shridhar2020alfworld}). AgentBench~\cite{liu2023agentbench} and AgentGym~\cite{xi2023agentgym} aggregate many environments under one evaluation protocol.

Newer benchmarks stress harder browsing, longer interaction, or richer diagnostics, including BrowseComp~\cite{wei2025browsecomp}, MIRAGE-Bench~\cite{zhang2025mirage}, holistic leaderboards~\cite{kapoor2025holistic}, and multi-turn RL with AgentGym-RL~\cite{xi2025agentgym}. The emphasis in most of this work is still whether the task is solved at the end. We instead sweep intrinsic horizon \(H^*\) and ask how the \emph{mix} of failure types changes as the horizon grows.

\subsection{Training, Planning, and Optimization}
\label{subsec:rw_training}

Agents are trained with reinforcement learning~\cite{ouyang2022training}, hierarchical planning~\cite{liu2023llmp,sun2023adaplanner}, tool use and reflection~\cite{yao2023react,shinn2023reflexion}, and deliberation-style prompting~\cite{wei2022chain,yao2023tree}. Recent papers also study planning on its own, for example plan--act pipelines for web agents~\cite{erdogan2025plan} and reliability of long plans in constrained settings~\cite{chen2024can}. That work improves how agents act. We study which failure modes become dominant as horizons grow, and what that implies for design (Section~\ref{sec:4}).

\subsection{Memory and Long-Context Reasoning}
\label{subsec:rw_memory}

Long context windows do not by themselves fix long-range reasoning~\cite{liu2023lost}. Better hardware and attention implementations~\cite{press2022train,dao2022flashattention} sit alongside ``lost in the middle'' and similar issues. Retrieval and external memory~\cite{madaan2022memory,lewis2020retrieval,park2023generative}, compression for trajectories~\cite{kang2025acon}, and learned memory--reasoning stacks~\cite{zhou2025mem1} tackle scaling from the architecture side. Embodied setups also maintain explicit state~\cite{wang2023voyager,ahn2022saycan}. We do not introduce a new memory module; we report how memory-related failures show up in labeled trajectories when we extend the horizon.

\subsection{Failure analysis and diagnostics}

Existing failure studies largely focus on domain-specific issues, such as code-generation errors~\cite{chen2023teaching}, web interaction failures~\cite{koh2024visualwebarena}, and embodied stochasticity~\cite{zeng2020tossingbot}.
Broader agent benchmarks can provide coarse failure categories: AgentBench~\cite{liu2023agentbench}, for example, reports high-level tags such as hallucination and planning errors, but does not study how failure patterns shift as task horizon increases.
Recent taxonomy-oriented work on multi-agent LLM systems~\cite{cemri2025multi} offers complementary diagnostics, yet its target setting is multi-agent coordination rather than single-agent, horizon-conditioned failures across domains.
We instead develop a cross-domain taxonomy grounded in FMEA and a structured literature review (Appendix~\ref{app:lit_mapping}), and validate trajectory-level labels with humans and a trajectory-grounded LLM-as-a-Judge pipeline (inter-annotator \(\kappa{=}0.61\), human--judge \(\kappa{=}0.84\); Appendix~\ref{app:4}).

\section{Appendix: Failure Taxonomy Definitions} \label{app:1}

\subsection{Environment Error}
\begin{enumerate}
\item \textbf{Environment
Disturbance / Problem Itself}

Environment failures arise when \textbf{the external world in which an agent operates changes,
degrades, or behaves stochastically in ways that are not explicitly modeled by the agent’s
plan or assumptions.} These disturbances include network or server instability, delayed or
partial UI rendering, unexpected pop-ups or overlays, schema and data drift, protocol
mismatches, and stochastic physical or simulator dynamics. Even when the agent’s
reasoning and action selection are locally correct, such environment-level disruptions can
invalidate trajectories by altering the state, interface, or action semantics mid-execution.

In long-horizon tasks, agents interact with the environment over extended time spans,
increasing the probability that the environment changes between planning, perception, and
execution. Cached assumptions (e.g., UI state, database schema, metrics, or action protocols)
become stale, and small disturbances compound across steps, leading to cascading failures
rather than immediate errors. As a result, long-horizon agents are particularly vulnerable
unless they incorporate robust environment re-validation, recovery policies (retry, refresh,
backoff), and explicit mechanisms to detect and adapt to environmental drift and noise.

\item \textbf{Unable to Detect
Environment Change (limitations
within the agent’s interaction
mechanisms)}

This failure occurs when an \textbf{agent cannot reliably perceive, verify, or internalize a change (or
non-change) in the environment, so its internal belief state diverges from the actual state.} It
often manifests as false transitions (believing a navigation or command succeeded when it
did not) or missed transitions (failing to notice the UI/OS/physical state has changed), driven
by interaction bottlenecks such as limited observability, imperfect state parsing, unsupported
action primitives (e.g., drag), or inability to handle obstacles like CAPTCHAs, banners, and
overlays. The result is that subsequent actions are executed against the wrong target state
(e.g., typing into the wrong page, issuing follow-up shell commands after a failed command,
or proceeding in manipulation despite a misaligned object placement).

Long-horizon tasks amplify this failure because state tracking must remain consistent across
many steps, and even a single undetected mismatch early on can cascade into repeated
invalid actions and compounding errors. Over longer trajectories, environments also present
more opportunities for partial loads, transient failures, UI changes, and evaluation-sensitive
deviations (e.g., “almost correct” placements), increasing the chance that the agent’s
interaction mechanism fails to register what actually happened. Without frequent re-
observation and state re-validation, the agent’s plan becomes increasingly decoupled from
reality, making recovery progressively harder.
\end{enumerate}
\input{fig_teaser}
\subsubsection{Web}
This failure occurs when a web agent operates under non-stationary or imperfectly perceived environments, including stochastic network conditions, dynamic or partially loaded pages, pop-ups/CAPTCHAs, or mismatches between the agent’s belief about the UI state and the actual rendered interface. In long-horizon web tasks, prolonged interaction increases the likelihood of such disturbances and belief–observation misalignment, causing agents to act on incomplete, outdated, or falsely assumed page states.

\subsubsection{OS}
This failure occurs when an OS agent operates under unexpected system-level conditions or misperceives command outcomes, such as permission denials, silent failures, or environment state changes that are not correctly reflected in the agent’s belief state. In long-horizon OS tasks, repeated command execution increases the likelihood that unobserved failures accumulate and desynchronize the agent’s internal state from the actual system state.

\subsubsection{Embodied} This failure occurs when an embodied agent’s execution is disrupted by external stochasticity or physical/environmental noise that invalidates previously correct plans or actions, or when the agent fails to detect such changes in the physical state. In long-horizon embodied tasks, prolonged interaction increases exposure to randomness (e.g., slippage, drops, perturbations), making environment-induced misalignment increasingly likely.

\subsubsection{Database}
This failure occurs when a database agent operates under non-stationary system environments, such as schema drift, model version updates, telemetry format changes, or external processes modifying the database state without the agent’s awareness. In long-horizon DB workflows, delayed execution and cached assumptions make agents especially vulnerable to acting on stale schemas or corrupted metrics.

\subsection{Instruction Error}
\begin{enumerate}
\item \textbf{Ill-defined
Instruction}

Ill-defined instruction failures occur when \textbf{the task specification is ambiguous, underspecified,
internally inconsistent, or misaligned with the capabilities and constraints of the operating
environment, making the intended goal state unclear or unreachable. }
This includes
unachievable goals (e.g., requesting non-existent fields or actions), out-of-scope follow-up
queries where required information is missing, and implicit or vague constraints (“safe,”
“active users,” “churn”) that are not formally defined. In such cases, agents may hallucinate
missing details, adopt incorrect default interpretations, or violate negative or safety
constraints while believing they are following the instruction.

Long-horizon tasks decompose high-level goals into many interdependent sub-instructions,
multiplying opportunities for ambiguity, drift, and constraint loss as execution unfolds. Early
misunderstandings or failed sub-steps propagate forward, causing later actions to be built on
invalid assumptions rather than triggering clarification or replanning. As trajectories grow
longer, the gap between the user’s original intent and the agent’s evolving interpretation
widens, making ill-defined instructions a dominant source of cascading, hard-to-detect failures.

\item \textbf{Partially Understand
Instruction}

This failure occurs when an agent grasps the high-level intent or primary action of an
instruction but fails to fully internalize critical constraints, modifiers, exceptions, or conditional
clauses. As a result, the agent performs actions that are syntactically valid and seemingly
reasonable, yet semantically incorrect with respect to the full instruction (e.g., ignoring
“except,” “only if,” or key qualifiers like “mismatching”). The agent may also misread
interaction feedback, prematurely concluding success despite visible error signals, warnings,
or unmet conditions in the execution trace.

In long-horizon tasks, instructions are often decomposed and re-represented across many
steps, increasing the risk that fine-grained constraints are dropped, weakened, or forgotten
over time. Partial understanding early in the trajectory can appear locally successful, masking
the error until much later when accumulated effects surface and recovery becomes costly.
Because long trajectories reward forward progress, agents may over-trust partial completion
signals and continue execution without re-checking whether all instruction conditions were
actually satisfied.
\end{enumerate}

\subsubsection{Web}
This failure arises when a web agent is given ill-defined, unachievable, or out-of-scope instructions that are inconsistent with the actual web environment or lack sufficient information to determine a valid goal state. In long-horizon workflows, instructions are decomposed into many dependent sub-requests, increasing the chance that ambiguities or unreachable goals propagate forward and cause cascading failures.

\subsubsection{OS}
Instruction failures arise when an OS agent understands the general task but misses critical constraints, exceptions, or negative instructions embedded in the prompt. In long-horizon settings, such omissions often propagate across multiple shell operations, leading to globally incorrect system modifications.

\subsubsection{Embodied} Instruction failures arise when an embodied agent captures only part of a task description but ignores critical modifiers or constraints, leading to correct actions applied to incorrect targets. Such partial understanding becomes more damaging in long-horizon tasks where early misinterpretations propagate across many physical steps.

\subsubsection{Database}
Instruction failures arise when a database agent misinterprets ambiguous, underspecified, or context-dependent natural-language requests, failing to align them with the correct business semantics or safety constraints. In long-horizon interactions, such misunderstandings propagate across planning and execution phases.

\subsection{Catastrophic Forgetting}
Catastrophic forgetting refers to the agent’s \textbf{failure to retain and consistently apply previously
acquired knowledge, constraints, or decisions while continuing to operate or adapt over time.}
This includes forgetting explicit user instructions (e.g., “never modify production” or “exclude
the payments schema”), earlier clarifications, rejected hypotheses, or prerequisite skills
needed for later subtasks. The failure can arise from limited context windows, inadequate
memory retrieval, or interference between newly introduced information and earlier task-
relevant state, causing the agent to revert to generic assumptions or unsafe defaults.

Long-horizon tasks inherently exceed the agent’s short-term context capacity and require
stable knowledge persistence across many turns, phases, or sessions. As execution
progresses, early constraints and learned insights are more likely to scroll out of context or be
overshadowed by recent interactions, leading to overwriting rather than accumulation of
knowledge. This results in cascading errors, repeated mistakes, and breakdowns in
dependency chains, where later steps silently violate earlier commitments or prerequisites that
are no longer actively represented.
\subsubsection{Web}
Catastrophic forgetting occurs when a web agent loses or stops enforcing earlier task-critical constraints (e.g., “new only,” “no spoilers,” “use main branch”) as the interaction progresses, often because those constraints drift out of active context or are overwritten by later subgoals.

\subsubsection{OS}
Catastrophic forgetting occurs when an OS agent loses earlier task constraints, safety requirements, or intermediate conclusions over the course of a multi-step execution. As OS tasks grow longer, earlier safeguards can drift out of active context, leading to irreversible or unsafe actions.

\subsubsection{Embodied} Catastrophic forgetting occurs when an embodied agent fails to retain prerequisite skills or learned environmental knowledge required for later stages of a long task. This is especially pronounced in hierarchical or curriculum-like tasks where later success depends on reliably transferring earlier competencies.

\subsubsection{Database}
Catastrophic forgetting occurs when a database agent loses earlier constraints, preferences, or hypotheses over extended multi-turn workflows, even though they remain relevant to later decisions. This leads to unsafe or inconsistent actions as long-horizon execution exceeds the agent’s effective retention capacity.

\subsection{False Assumption}
\begin{enumerate}
\item \textbf{Universal
Facts (Overgeneralized World
Knowledge)}

This failure occurs when an \textbf{agent relies on implicit assumptions, heuristics, or patterns learned from prior experience and treats them as universally true, without verifying whether they hold in the current task or environment.} Unlike environment failures—where the external world changes or behaves unpredictably—this error arises from the agent’s incorrect prior belief about how the environment should behave, leading it to generate actions or reasoning based on a false premise.
These false assumptions can include hallucinated APIs or schema elements, default interpretations of goals, over-trusting probabilistic components as deterministic, or inferring causality from correlation. Even when local reasoning steps appear coherent, the agent’s trajectory is grounded in an incorrect premise, causing repeated ineffective actions or confidently wrong decisions.

Long-horizon reasoning encourages agents to commit early to implicit assumptions and reuse them across many steps, making it less likely that those assumptions are revisited or challenged later. As trajectories grow longer, confirmation bias sets in: intermediate outputs are interpreted as validation rather than signals to re-check premises, allowing false assumptions to silently propagate and compound. Without explicit mechanisms for assumption verification or hypothesis revision, long-horizon agents increasingly drift from the true environment dynamics while remaining internally consistent.

\item \textbf{Long-
Horizon Complex Task Reasoning}

This failure arises when an agent implicitly assumes it can sustain correct multi-step logic, control flow, and state consistency across long and complex task sequences, but in practice loses coherence as execution unfolds. It manifests as incorrect loop or conditional logic, premature termination, infinite repetition, or divergence mid-trajectory due to unverified assumptions about system state (e.g., file ownership, working directory, or intermediate progress). Although individual actions may be locally reasonable, the agent’s global reasoning chain breaks because earlier assumptions are neither validated nor revised.

Long-horizon tasks require maintaining structured plans, intermediate invariants, and progress tracking over many dependent steps, which standard agents often approximate with shallow or implicit reasoning. As task length increases, small logical errors or unchecked state assumptions compound, causing the agent to drift off-goal or exhaust interaction budgets without recovery. Without explicit planning, loop verification, and state re-checking, agents increasingly rely on false assumptions of coherence, making complex long-horizon tasks a primary failure regime.

\end{enumerate}
\subsubsection{Web}
This failure occurs when a web agent fills in missing information or applies generic priors as if they were universally valid, instead of grounding decisions in what is actually observable on the webpage (e.g., fabricating an email, assuming a policy is allowed, or treating ambiguity as resolved).
\subsubsection{OS}
This failure occurs when an OS agent assumes system properties or command semantics without verification, relying on generic priors instead of observed system state. Such assumptions can lead to incorrect branching decisions or invalid operations, even when commands are syntactically correct.

\subsubsection{Embodied}
This failure occurs when an embodied agent overgeneralizes from common training patterns, assuming that a visually similar scene implies a standard goal or action, regardless of the actual instruction. Such bias-driven assumptions override task-specific constraints and lead to systematically wrong behavior.

\subsubsection{Database}
This failure occurs when a database agent overgeneralizes schema patterns, operator semantics, or workload behavior, treating them as universally valid without verification. Such assumptions override actual database-specific details and lead to systematically incorrect reasoning.

\subsection{Planning Error}
\begin{enumerate}
\item \textbf{Sub-plan.}
Sub-plan planning errors occur when an \textbf{agent mis-decomposes a high-level goal into incorrect, incomplete, poorly ordered, or inefficient sub-steps, even though the overall objective is understood.} This includes choosing the wrong API or parameter format for a specific step, omitting prerequisite actions, violating constraints within a sub-plan, or adopting an inefficient strategy that exhausts time, memory, or interaction budgets. The failure is localized at the sub-plan level, but it breaks the global task by introducing deadlocks, irreversible actions, or cascading inefficiencies.

Long-horizon tasks magnify sub-plan errors because later steps critically depend on the correctness and ordering of earlier ones, and recovery options shrink as execution progresses. A single flawed sub-plan (e.g., dropping backups too early, rebuilding indexes before diagnosis) can invalidate all downstream actions or make rollback difficult. As plans grow deeper and more branched, agents without explicit plan validation or cost-aware reasoning increasingly commit to locally plausible but globally harmful sub-plans, leading to systematic long-horizon failure.
\item \textbf{Action.}
Action-level planning errors occur when an \textbf{agent selects or executes an incorrect concrete action, despite having an appropriate high-level plan or sub-task in place.} These failures include execution crashes (e.g., syntax or type errors), semantically invalid actions that do not achieve the intended effect in the environment, or mismatches between planned intent and low-level action realization (e.g., incorrect API flags, shell commands, or motor control tokens). As a result, the action either fails outright or silently alters the environment in unintended ways.

In long-horizon tasks, agents must execute many interdependent actions, making even small action-level errors costly because they propagate forward and constrain future options. Repeated low-level failures can exhaust interaction budgets, corrupt intermediate state, or trigger compounding side effects before the agent detects the problem. Without robust action validation, rollback, and feedback-aware correction, long-horizon execution amplifies these errors from isolated glitches into trajectory-ending failures.
\end{enumerate}

\subsubsection{Web}
Planning errors occur when a web agent chooses an incorrect sequence of subgoals or misses key prerequisite steps, leading it to pursue the wrong navigation branch, ignore constraints, or stop prematurely even if the end goal is clear.
\subsubsection{OS}
Planning errors arise when an OS agent fails to decompose a high-level instruction into a correct and ordered sequence of shell operations, leading to missing steps, incorrect ordering, or inappropriate operator selection. These errors often surface before execution, but their effects compound over long execution chains.

\subsubsection{Embodied}
Planning errors arise when an embodied agent generates an incorrect sequence of sub-goals or fails to integrate multi-step logical requirements, even though individual actions may be physically executable. In long-horizon settings, weak decomposition or missing logical dependencies leads to deadlocks or goal divergence.

\subsubsection{Database}
Planning errors arise when a database agent constructs an incorrect or suboptimal sequence of logical or operational steps, even if each individual operation is valid in isolation. In long-horizon DB tasks, poor plan ordering or granularity can cause irreversible damage or severe inefficiency.

\subsection{History Error Accumulation}
History error accumulation refers to failures where \textbf{small mistakes, incorrect assumptions, or mis-specified intermediate artifacts introduced early in a trajectory persist in the agent’s working state and progressively distort downstream reasoning and actions}. The “history” can include intermediate tables, cached schemas, partial code edits, stored hypotheses, retrieved documents, or planning traces. Because later steps reuse these artifacts as if they were correct, the system can produce outputs that are internally consistent yet fundamentally wrong (e.g., subtle over-counting, reasoning over the wrong feature column, or refining a plan built on a misdiagnosis).

Long-horizon workflows amplify this failure because they involve many dependent iterations, where each step inherits context and intermediate products from previous steps. Even a minor upstream error (a missing filter, wrong ID, biased retrieval) can silently propagate across multiple tools, agents, and refinements, making the final failure difficult to localize and correct. As trajectories lengthen and artifacts become more complex (e.g., long SQL scripts or multi-stage playbooks), agents increasingly apply “patchy” local edits instead of globally re-validating assumptions, allowing errors to compound into trajectory-level breakdowns.
\subsubsection{Web}
History error accumulation occurs when a web agent fails to correctly interpret or update its interaction history, causing small mistakes or ineffective actions to be repeated and compounded over time rather than corrected. In long-horizon web tasks, such uncorrected repetitions (e.g., reusing stale element IDs, appending corrupted queries, or toggling controls) can exponentially degrade the agent’s state and exhaust the interaction budget.

\subsubsection{OS}
History error accumulation occurs when errors or misinterpretations in early command executions are not corrected, and the agent continues reasoning as if those steps succeeded. Over long OS interaction trajectories, such uncorrected assumptions compound and increasingly diverge from the true system state.

\subsubsection{Embodied}
History error accumulation occurs when an embodied agent fails to correct an early planning or execution error and continues acting based on that flawed history, compounding the failure over time. In long-horizon tasks, this often results in persistent loops or deadlocks instead of replanning.

\subsubsection{Database}
History error accumulation occurs when early logical or data-level mistakes are reused by downstream steps, causing subtle errors to compound across queries, plans, or optimizations. Because later stages treat earlier artifacts as ground truth, the system becomes increasingly confident yet increasingly wrong.

\subsection{Memory Limitations}
Memory limitation failures occur when an agent cannot retain, retrieve, or reason over all task-relevant information because \textbf{the interaction history, environment state, or intermediate artifacts exceed its effective memory capacity}. This includes exceeding the model’s context window, lossy summarization of prior steps, or selective retrieval that omits critical constraints, schema elements, or past observations. As a result, the agent may operate with an incomplete or distorted view of the task, leading to inconsistent joins, dropped filters, unsafe recommendations, or misinterpretation of system state.

Long-horizon tasks inherently accumulate large amounts of state—schemas, logs, plans, hypotheses, and prior decisions—quickly overwhelming fixed-context or shallow memory mechanisms. As execution continues, earlier but still critical facts fall out of context or are compressed into coarse summaries, making subtle edge cases and historical dependencies invisible to later reasoning. Without robust external memory, selective retrieval, and consistency checks, long-horizon agents increasingly behave as if earlier constraints or observations never existed, causing systematic drift and brittle decision-making.
\subsubsection{Web}
Memory limitation failures arise when a web agent cannot retain and compare multi-item evidence across steps (due to context limits or lack of external scratchpad), causing it to drop earlier observations, constraints, or candidate details.
\subsubsection{OS}
Memory limitation failures arise when an OS agent cannot retain earlier command outputs, constraints, or environment details because the interaction history exceeds its effective context capacity. In long OS tasks, this leads to violations of earlier constraints or repeated mistakes due to missing historical information.

\subsubsection{Embodied}
Memory limitation failures arise when an embodied agent cannot maintain or retrieve sufficient state information over extended horizons, such as object locations, intermediate goals, or past outcomes. As task length increases, limited internal memory leads to inconsistent behavior and loss of progress tracking.

\subsubsection{Database}
Memory limitation failures arise when a database agent cannot keep large schemas, long interaction histories, or complex multi-modal plans fully in context, forcing lossy summarization or selective retrieval. Over long horizons, this results in dropped constraints, missing joins, or inconsistent reasoning across turns.

\clearpage
\section{Appendix: Literature-to-Taxonomy Mapping} \label{app:lit_mapping}

We systematically reviewed long-horizon agent literature using a shared spreadsheet (\texttt{Literature Review\_Long-Horizon Reasoning in LLM agents.xlsx}). For each paper we recorded title, venue, domains, models, and free-text notes on reported failure mechanisms (the spreadsheet column \emph{Failure Modes}). We then aligned recurring mechanisms to our seven-category taxonomy during iterative coding (Section~\ref{subsec:failure_taxonomy}). Table~\ref{tab:lit_taxonomy_map} lists \emph{three representative prior works per category} drawn from that review corpus and already present in our bibliography; these are illustrative anchors rather than an exhaustive list.

\begin{table*}[h]
\centering
\small
\caption{\textit{Representative prior work mapped to each failure category.} Entries are drawn from our literature spreadsheet and selected for cross-domain coverage; full paper lists and extracted notes remain in the supplementary spreadsheet.}
\label{tab:lit_taxonomy_map}
\resizebox{\textwidth}{!}{%
\begin{tabular}{@{}p{2.6cm}p{9.2cm}p{4.6cm}@{}}
\toprule
\textbf{Category} & \textbf{Representative prior works (3)} & \textbf{Typical mechanism (from review notes)} \\
\midrule
Environment &
\cite{zhou2023webarena,chen2025robohorizon,koh2024visualwebarena} &
Environment fidelity, transitions, or multimodal observation mismatch. \\
Instruction &
\cite{trivedi2024appworld,chen2024can,wang2025odysseybench} &
Ambiguous goals, missing constraints, or mis-specified task requirements. \\
Catastrophic Forgetting &
\cite{sinha2025illusion,li2024larm,fengevoagent} &
Loss of long-range constraints or competencies as trajectories lengthen. \\
False Assumption &
\cite{zhang2025mirage,yang2024agentoccam,wei2025browsecomp} &
Hallucinated facts, unfounded shortcuts, or misaligned beliefs about state. \\
Planning Error &
\cite{erdogan2025plan,si2025goal,gonzalez2025robotouille} &
Subgoal decomposition, ordering, or plan repair under dynamics. \\
History Error Accumulation &
\cite{wang2025ragen,liu2023agentbench,xi2025agentgym} &
Multi-step error propagation, unstable rollouts, or interaction loops. \\
Memory Limitation &
\cite{kang2025acon,zhou2025mem1,anwar2025remembr} &
Context limits, compression trade-offs, or long-horizon memory bottlenecks. \\
\bottomrule
\end{tabular}}
\end{table*}

\clearpage
\section{Appendix: Experiment Set-up} \label{app:2}
Although a universal breaking point is unattainable, empirically the notion of a breaking point remains meaningful as a model- and domain-conditional quantity. Different agents exhibit markedly different long-horizon capacities, and the same agent may fail at very different horizons depending on domain characteristics such as environment dynamics, observability, and action semantics.
Operationally, we treat a breaking point not as a single universal threshold, but as a transition region on a performance–horizon curve, where success rates begin to collapse sharply and failures shift from recoverable local errors to irreversible trajectory derailment.

Using our proposed \texttt{HORIZON} framework, we demonstrate the feasibility of systematically approximating such breaking behavior across domains. Our goal is to validate \texttt{HORIZON} as aan initial cross-domain diagnostic benchmark that enables controlled horizon extension and comparable analysis along both the horizontal (horizon growth) and vertical (failure attribution) dimensions, serving as an initial step toward principled long-horizon research for agentic systems.

\subsection{Data, Environments, and Agents}
Given the novelty of our horizon definition and diagnostic framework, no existing benchmark directly supports the proposed notion of intrinsic horizon and controlled extension. We therefore construct our evaluation data using the \texttt{HORIZON} workflow illustrated in Figure~\ref{fig:workflow}. Across all domains, baseline tasks are first verified to be solvable, ensuring that performance degradation arises from increased horizon rather than task ambiguity or agent misconfiguration.
\begin{enumerate}[itemsep=0pt, topsep=2pt, parsep=0pt]
\item \textit{Web Navigation}. We adopt WebArena~\cite{zhou2023webarena} as the base environment. From the benchmark task pool, we select tasks that achieve 100\% success under a baseline run with GPT-5-mini, and apply our horizon extension procedures to construct task families with increasing intrinsic horizon.
\item \textit{Operating Systems (OS)}. We build on OS tasks from AgentBench~\cite{liu2023agentbench}, again selecting tasks with perfect baseline performance under GPT-5-mini before applying depth-based horizon extension.
\item \textit{Databases (DB)}. We build on MAC-SQL~\cite{wang2025macsqlmultiagentcollaborativeframework}, a multi-agent Text-to-SQL framework, to support controlled query decomposition and task composition.
The agent follows a modular pipeline consisting of three components: a \textit{Selector}, a \textit{Decomposer}, and a \textit{Refiner}. Given a user query, the Selector first identifies relevant tables and schema elements. The Decomposer then breaks the query into sub-questions and generates intermediate SQL statements, which are composed into a final query. Finally, the Refiner executes the generated SQL, detects errors (e.g., execution or syntax errors), and iteratively revises the query to produce the final result.

\item \textit{Embodied}. We developed a bimanual robot arm simulation environment in IsaacSim 5.0 \cite{kachaev2025mind}. This environment consisted of two Franka Emika Panda Robot Arms, each equipped with a Tesollo DG-3F-B three-finger gripper. To evaluate variable-horizon tasks in an embodied setting, we developed a bimanual robot arm simulation environment using IsaacSim~5.0. The environment consists of two Franka Emika Panda robot arms, each equipped with a Tesollo DG-3F-B three-finger gripper.

We define four primitive robot actions, each parameterized and directly controllable by the agent:
\begin{itemize}[leftmargin=*, itemsep=2pt]
    \item \texttt{home}: Move the entire robot system to its default configuration.
    \item \texttt{move\_to\_pose}: Move a specified gripper to a target Cartesian pose (position and orientation).
    \begin{itemize}[leftmargin=2em, itemsep=1pt]
        \item \textbf{Parameters:}
        \begin{description}[leftmargin=2em, style=nextline]
            \item[\texttt{arm}] The arm to move (\texttt{left} or \texttt{right}).
            \item[\texttt{pose}] The target position and orientation.
        \end{description}
    \end{itemize}
    \item \texttt{grasp}: Close the gripper.
    \begin{itemize}[leftmargin=2em, itemsep=1pt]
        \item \textbf{Parameters:}
        \begin{description}[leftmargin=2em, style=nextline]
            \item[\texttt{arm}] The gripper to close (\texttt{left} or \texttt{right}).
        \end{description}
    \end{itemize}
    \item \texttt{release}: Open the gripper.
    \begin{itemize}[leftmargin=2em, itemsep=1pt]
        \item \textbf{Parameters:}
        \begin{description}[leftmargin=2em, style=nextline]
            \item[\texttt{arm}] The gripper to open (\texttt{left} or \texttt{right}).
        \end{description}
    \end{itemize}
\end{itemize}

The simulation environment is populated with three objects: a red cube, a blue cube, and a green cube. The available actions and object locations are provided to the agent as part of the task context. For each task, the agent is prompted to generate a JSON-formatted plan composed solely of the predefined actions and their parameters. The generated plan is then executed in simulation, and task success is evaluated manually. 

\end{enumerate}

\subsection{Models} 
To demonstrate feasibility on current frontier systems, we evaluate two state-of-the-art representative model families. 
We use GPT-5-mini \cite{openai-gpt5-mini} and Claude-4-Sonnet \cite{anthropic-claude4-sonnet} across all four domains.

\subsection{Evaluation}
We evaluate agent performance as a function of the \texttt{HORIZON} extension level $s$, reporting success rates over task sets with identical $s$ shown in Figure~\ref{fig:workflow}. In addition to outcome-level success, failed trajectories are analyzed using our seven-category failure taxonomy (Section~\ref{subsec:failure_taxonomy}) through a combination of expert human annotation and calibrated LLM-based judges. 

To quantify variability, we report mean and standard deviation over three independent runs for each horizon setting. In the Embodied domain, both model families show clear horizon-dependent degradation, with higher variance at intermediate horizons. For GPT-5 variants, success decreases from \(0.926 \pm 0.064\) at short horizon to \(0.407 \pm 0.170\), \(0.259 \pm 0.231\), and \(0.074 \pm 0.064\) as horizon increases. Claude-4 shows a similar trend, dropping from \(0.963 \pm 0.064\) to \(0.259 \pm 0.064\), \(0.111 \pm 0.111\), and \(0.037 \pm 0.064\). 

\clearpage
\section{Appendix: 5-domain $\times$ 7-type Failure Mini-Trajectories} \label{app:3}
\definecolor{custompink}{HTML}{DD99BB} 
\definecolor{customred}{HTML}{F66B2E} 
\definecolor{customyellow}{HTML}{F3AE36} 
\definecolor{customgreen}{HTML}{B9D5B1} 
\definecolor{customblue}{HTML}{4D9DE0} 
\definecolor{custompurple}{HTML}{AEADF0} 
\definecolor{customwhite}{HTML}{fbf9f4}

\subsection{Web}
\begin{tcolorbox}[
breakable,
colback=customwhite,
colframe=custompurple,
title=Environment,
fontupper=\fontfamily{pcr}\selectfont\scriptsize
]

\textbf{\textcolor{red}{Step 1 (FAILURE STEP)}}

\textbf{Summary:}  
The agent attempts to go to the homepage and issues a goto that resolves to localhost:4399.

\textbf{Raw Trace:}
\begin{verbatim}
Page : Tab 0 (current): One Stop Market
Agent: I need to analyze this objective which has three parts:
1. What do customers say about brush from sephora
2. Tell me the total cost of my latest cancelled order
3. Tell me the total cost of my latest pending order

I'm currently on the One Stop Market homepage, but I need to go to
Sephora for the first part of the objective. Let me navigate to the
homepage to find Sephora.

In summary, the next action I will perform is
goto [http://homepage.com]

Action: goto [localhost:4399]
\end{verbatim}

\medskip
\textbf{Step 2}

\textbf{Summary:}  
Seeing a Chrome error page, the agent retries navigating to the homepage but again executes goto localhost:4399.

\textbf{Raw Trace:}
\begin{verbatim}
Page : Tab 0 (current):
Agent: I can see that the previous action tried to navigate to
localhost:4399 but resulted in a Chrome error page. This suggests
the local server might not be running or the URL is incorrect.
Let me try to navigate to the homepage to find available websites.

In summary, the next action I will perform is
goto [http://homepage.com]

Action: goto [localhost:4399]
\end{verbatim}

\medskip
\textbf{Step 3}

\textbf{Summary:}  
The agent notes the error again and repeats the same navigation action to localhost:4399.

\textbf{Raw Trace:}
\begin{verbatim}
Page : Tab 0 (current):
Agent: I can see that the current page is showing a Chrome error
page, which means the URL I tried to navigate to (localhost:4399)
is not accessible. Let me try to navigate to the homepage to see
what websites are available.

In summary, the next action I will perform is
goto [http://homepage.com]

Action: goto [localhost:4399]
\end{verbatim}

\end{tcolorbox}
\clearpage

\begin{tcolorbox}[
breakable,
colback=customwhite,
colframe=custompurple,
title=Instruction,
fontupper=\fontfamily{pcr}\selectfont\scriptsize
]

\textbf{Step 1}

\textbf{Summary:}  
On the order page, the agent clicks the Cancel button.

\textbf{Raw Trace:}
\begin{verbatim}
[Step 8]
Page : Tab 0 (current): #000000307 / Orders / Operations / Sales / Magento Admin
Agent: In summary, the next action I will perform is
click [13783]

Action: click [13783] where [13783] is button 'Cancel'
\end{verbatim}

\medskip
\textbf{Step 2}

\textbf{Summary:}  
The agent confirms the cancellation by clicking OK.

\textbf{Raw Trace:}
\begin{verbatim}
[Step 9]
Page : Tab 0 (current): #000000307 / Orders / Operations / Sales / Magento Admin
Agent: In summary, the next action I will perform is
click [16102]

Action: click [16102] where [16102] is button 'OK' focused: True
\end{verbatim}

\medskip
\textbf{\textcolor{red}{Step 3 (FAILURE STEP)}}

\textbf{Summary:}  
The agent stops with a narrative message instead of the required single-word answer "Canceled".

\textbf{Raw Trace:}
\begin{verbatim}
[Step 10]
Page : Tab 0 (current): #000000307 / Orders / Operations / Sales / Magento Admin
Agent: In summary, the next action I will perform is
stop [Order #000000307 is already canceled]

Action: stop [Order #000000307 is already canceled]
\end{verbatim}

\end{tcolorbox}

\clearpage

\begin{tcolorbox}[
breakable,
colback=customwhite,
colframe=custompurple,
title=Catastrophic Forgetting,
fontupper=\fontfamily{pcr}\selectfont\scriptsize
]

\textbf{Step 1}

\textbf{Summary:}  
The agent clicks the first product's "Add to Wish List" button from the laundry detergent results.

\textbf{Raw Trace:}
\begin{verbatim}
[Step 3]
Page : Tab 0 (current): Search results for: 'laundry detergent'
Agent: Let me add the first laundry detergent (Tide PODS)
to my wish list.

Action: click [5136] where [5136] is button 'Add to Wish List'
\end{verbatim}

\medskip
\textbf{Step 2}

\textbf{Summary:}  
On the Wish List page, the agent confirms the detergent was added and navigates to My Orders.

\textbf{Raw Trace:}
\begin{verbatim}
[Step 4]
Page : Tab 0 (current): My Wish List

Agent: The detergent has been added to the wish list.
Now I need to check my latest cancelled order.

Action: click [7019] where [7019] is link 'My Orders'
\end{verbatim}

\medskip
\textbf{\textcolor{red}{Step 3 (FAILURE STEP)}}

\textbf{Summary:}  
Despite reading the latest cancelled order and its total ($\$365.42$), the agent wrongly decides it still needs to add a detergent.

\textbf{Raw Trace:}
\begin{verbatim}
[Step 5]
Page : Tab 0 (current): My Orders

Agent: The latest cancelled order is #000000170
with a total of $365.42.

Now I need to add a laundry detergent to my wish list.

Action: click [9227] menuitem 'Home & Kitchen'
\end{verbatim}

\medskip
\textbf{Step 4}

\textbf{Summary:}  
The agent navigates to Home \& Kitchen and starts searching again.

\textbf{Raw Trace:}
\begin{verbatim}
[Step 6]
Page : Tab 0 (current): Home & Kitchen

Action: type [12303] [laundry detergent]
\end{verbatim}

\medskip
\textbf{Step 5}

\textbf{Summary:}  
The agent scrolls the search results to find products.

\textbf{Raw Trace:}
\begin{verbatim}
[Step 7]
Page : Tab 0 (current): Search results for: 'laundry detergent'

Action: scroll [down]
\end{verbatim}

\end{tcolorbox}

\begin{tcolorbox}[
breakable,
colback=customwhite,
colframe=custompurple,
title=False Assumption,
fontupper=\fontfamily{pcr}\selectfont\scriptsize
]

\textbf{Step 1}

\textbf{Summary:}  
On the Customer Reviews Report, the agent clicks "Show Reviews".

\textbf{Raw Trace:}
\begin{verbatim}
[Step 4]
Page : Customer Reviews Report / Magento Admin

Action: click [2068] link 'Show Reviews'
\end{verbatim}

\medskip
\textbf{Step 2}

\textbf{Summary:}  
The agent observes one customer with one review but switches to the "By Products" report.

\textbf{Raw Trace:}
\begin{verbatim}
[Step 5]
Page : Customer Reviews Report

Agent: Emma Lopez has 1 review.
Let me check the "By Products" report.

Action: click [1611] link 'By Products'
\end{verbatim}

\medskip
\textbf{\textcolor{red}{Step 3 (FAILURE STEP)}}

\textbf{Summary:}  
The agent misinterprets "126 records found" as the total number of reviews.

\textbf{Raw Trace:}
\begin{verbatim}
[Step 6]
Page : Product Reviews Report

Agent: The page shows "126 records found",
which indicates the total number of reviews.

Action: stop [126]
\end{verbatim}

\end{tcolorbox}

\begin{tcolorbox}[
breakable,
colback=customwhite,
colframe=custompurple,
title=Planning Error,
fontupper=\fontfamily{pcr}\selectfont\scriptsize
]

\textbf{Step 1}

\textbf{Summary:}  
From the Bestsellers report page, the agent opens the Customers section.

\textbf{Raw Trace:}
\begin{verbatim}
Page : Bestsellers Report / Magento Admin

Action: click [5882] link 'CUSTOMERS'
\end{verbatim}

\medskip
\textbf{Step 2}

\textbf{Summary:}  
The agent opens the "All Customers" page.

\textbf{Raw Trace:}
\begin{verbatim}
Page : Bestsellers Report

Action: click [6286] link 'All Customers'
\end{verbatim}

\medskip
\textbf{\textcolor{red}{Step 3 (FAILURE STEP)}}

\textbf{Summary:}  
Instead of displaying the customer grid, the agent attempts to export the customer list.

\textbf{Raw Trace:}
\begin{verbatim}
Page : Customers / Magento Admin

Agent: To show all customers, I will export the list.

Action: click [8519] button 'Export'
\end{verbatim}

\medskip
\textbf{Step 4}

\textbf{Summary:}  
The agent continues the export workflow.

\textbf{Raw Trace:}
\begin{verbatim}
Page : Customers / Magento Admin

Action: click [10946] button 'Export'
\end{verbatim}

\medskip
\textbf{Step 5}

\textbf{Summary:}  
The agent switches focus back to Reports, indicating a lack of coherent plan.

\textbf{Raw Trace:}
\begin{verbatim}
Page : Customers / Magento Admin

Action: click [6998] link 'REPORTS'
\end{verbatim}

\end{tcolorbox}

\begin{tcolorbox}[
breakable,
colback=customwhite,
colframe=custompurple,
title=History Error Accumulation,
fontupper=\fontfamily{pcr}\selectfont\scriptsize
]

\textbf{Step 1}

\textbf{Summary:}  
On the My Orders page, the agent prematurely navigates to the Video Games category instead of inspecting the orders to find the latest cancelled order total.

\textbf{Raw Trace:}
\begin{verbatim}
[Step 4]
Page : My Orders

Agent: I will open the Video Games category first.

Action: click [8961]  # Video Games
\end{verbatim}

\medskip
\textbf{Step 2}

\textbf{Summary:}  
After returning to the My Orders page later in the trajectory, the agent repeats the same navigation mistake and again leaves the page without checking the order information.

\textbf{Raw Trace:}
\begin{verbatim}
[Step 10]
Page : My Orders

Agent: I will open the Video Games category first.

Action: click [24391]  # Video Games
\end{verbatim}

\medskip
\textbf{Step 3}

\textbf{Summary:}  
The agent correctly identifies the cancelled order total ($\$365.42$) but still navigates away instead of reporting the answer, allowing the earlier mistake to persist.

\textbf{Raw Trace:}
\begin{verbatim}
[Step 32]
Page : My Orders

Agent: The latest cancelled order visible here is
order 000000170 dated 5/17/23 with total $365.42.
I will report it after navigating to Video Games.

Action: click [82213]  # Video Games
\end{verbatim}

\medskip
\textbf{\textcolor{red}{Step 4 (FAILURE STEP)}}

\textbf{Summary:}  
Because the agent repeatedly leaves the relevant page instead of reporting the answer, the trajectory eventually reaches the maximum step limit and terminates without completing the task.

\textbf{Raw Trace:}
\begin{verbatim}
[Step 61]
Page : Video Games

Agent:

Action: stop [Early stop: Reach max steps 60]
\end{verbatim}

\end{tcolorbox}

\begin{tcolorbox}[
breakable,
colback=customwhite,
colframe=custompurple,
title=History Error Accumulation,
fontupper=\fontfamily{pcr}\selectfont\scriptsize
]

\textbf{Step 1}

\textbf{Summary:}  
The agent counts case-insensitive occurrences of the word “error” in all log files and writes the results into \texttt{logs/file\_error\_counts.txt}.

\textbf{Raw Trace:}
\begin{verbatim}
[Step 1]

Agent: Count case-insensitive "error" occurrences in
each log file and write "filename: count" to
logs/file_error_counts.txt.

Action: bash
awk 'BEGIN{IGNORECASE=1} {n=gsub(/error/,"&");
counts[FILENAME]+=n}
END{for (f in counts) print f": "counts[f]}' \
logs/log_*.txt > logs/file_error_counts.txt
\end{verbatim}

\medskip
\textbf{Step 2}

\textbf{Summary:}  
The agent checks the number of lines in the generated file and observes 8 entries instead of the expected 7.

\textbf{Raw Trace:}
\begin{verbatim}
[Step 2]

Agent: Check that logs/file_error_counts.txt contains
seven lines.

Action: bash
wc -l logs/file_error_counts.txt
\end{verbatim}

\medskip
\textbf{Step 3}

\textbf{Summary:}  
To diagnose the discrepancy, the agent inspects the contents of the output file.

\textbf{Raw Trace:}
\begin{verbatim}
[Step 3]

Agent: Inspect the file contents to understand why
there are 8 lines instead of 7.

Action: bash
cat logs/file_error_counts.txt
\end{verbatim}

\medskip
\textbf{\textcolor{red}{Step 4 (FAILURE STEP)}}

\textbf{Summary:}  
Instead of continuing the debugging process, the agent abruptly executes an unrelated command to count files in \texttt{/etc}, indicating that the earlier reasoning steps have gradually drifted away from the original task objective.

\textbf{Raw Trace:}
\begin{verbatim}
[Step 4]

Agent: Count the number of regular files directly
in /etc.

Action: bash
find /etc -maxdepth 1 -type f | wc -l
\end{verbatim}

\end{tcolorbox}

\subsection{OS}
\begin{tcolorbox}[
breakable,
colback=customwhite,
colframe=customyellow,
title=Environment,
fontupper=\fontfamily{pcr}\selectfont\scriptsize
]

\textbf{Step 1}

\textbf{Summary:}  
Make the wrapper script executable using chmod.

\textbf{Raw Trace:}
\begin{verbatim}
chmod +x date-format
(OS response: empty)
\end{verbatim}

\medskip
\textbf{Step 2}

\textbf{Summary:}  
Create README.md using a heredoc containing usage instructions.

\textbf{Raw Trace:}
\begin{verbatim}
cat > README.md << 'EOF'
# Date Format Tool
...
EOF
\end{verbatim}

\medskip
\textbf{\textcolor{red}{Step 3 (FAILURE STEP)}}

\textbf{Summary:}  
Attempt to list directory contents but receive no output.

\textbf{Raw Trace:}
\begin{verbatim}
ls -la
(OS response: empty)
\end{verbatim}

\medskip
\textbf{Step 4}

\textbf{Summary:}  
Try to print the working directory but again receive empty output.

\textbf{Raw Trace:}
\begin{verbatim}
pwd
(OS response: empty)
\end{verbatim}

\medskip
\textbf{Step 5}

\textbf{Summary:}  
Run a simple echo command which also returns no output.

\textbf{Raw Trace:}
\begin{verbatim}
echo "test"
(OS response: empty)
\end{verbatim}

\end{tcolorbox}

\begin{tcolorbox}[
breakable,
colback=customwhite,
colframe=customyellow,
title=Instruction,
fontupper=\fontfamily{pcr}\selectfont\scriptsize
]

\textbf{Step 1}

\textbf{Summary:}  
Count lines in verification.log to check for discrepancies.

\textbf{Raw Trace:}
\begin{verbatim}
wc -l verification.log
\end{verbatim}

\medskip
\textbf{Step 2}

\textbf{Summary:}  
Generate \verb|final_prep.txt| reporting file and directory counts.

\textbf{Raw Trace:}
\begin{verbatim}
awk 'NR==1{f=$0} NR==2{d=$0}
END{printf "Verified file count: %s\nVerified directory count: %s\n",f,d}'
etc_counts.txt > final_prep.txt
\end{verbatim}

\medskip
\textbf{\textcolor{red}{Step 3 (FAILURE STEP)}}

\textbf{Summary:}  
Incorrectly answers using only the file count rather than the required total.

\textbf{Raw Trace:}
\begin{verbatim}
answer(There are 58 files in /etc.)
\end{verbatim}

\end{tcolorbox}

\clearpage

\begin{tcolorbox}[
breakable,
colback=customwhite,
colframe=customyellow,
title=Catastrophic Forgetting,
fontupper=\fontfamily{pcr}\selectfont\scriptsize
]

\textbf{Step 9}

\textbf{Summary:}  
Attempt a multiline awk script but quoting fails.

\textbf{Raw Trace:}
\begin{verbatim}
awk ' ... ' temp_file_list.txt > txt_sizes.csv
(OS response echoes script)
\end{verbatim}

\medskip
\textbf{Step 10}

\textbf{Summary:}  
Retry using a one-line awk command.

\textbf{Raw Trace:}
\begin{verbatim}
awk ' ... ' temp_file_list.txt > txt_sizes.csv
(OS response: empty)
\end{verbatim}

\medskip
\textbf{\textcolor{red}{Step 11 (FAILURE STEP)}}

\textbf{Summary:}  
Abruptly switches to counting files under /etc, abandoning the original task.

\textbf{Raw Trace:}
\begin{verbatim}
find /etc -type f | wc -l
-> 162
\end{verbatim}

\medskip
\textbf{Step 12}

\textbf{Summary:}  
Reports the /etc file count as the final answer.

\textbf{Raw Trace:}
\begin{verbatim}
answer(There are 162 regular files in /etc.)
\end{verbatim}

\end{tcolorbox}

\begin{tcolorbox}[
breakable,
colback=customwhite,
colframe=customyellow,
title=False Assumption,
fontupper=\fontfamily{pcr}\selectfont\scriptsize
]

\textbf{Step 1}

\textbf{Summary:}  
List regular files into \verb|file_list.txt|.

\textbf{Raw Trace:}
\begin{verbatim}
find . -maxdepth 1 -type f -printf '%P\n' > file_list.txt
(OS response: empty)
\end{verbatim}

\medskip
\textbf{Step 2}

\textbf{Summary:}  
Create the matches directory.

\textbf{Raw Trace:}
\begin{verbatim}
mkdir -p matches
(OS response: empty)
\end{verbatim}

\medskip
\textbf{\textcolor{red}{Step 3 (FAILURE STEP)}}

\textbf{Summary:}  
Incorrectly concludes that no files exist due to empty console output.

\textbf{Raw Trace:}
\begin{verbatim}
answer(0)
\end{verbatim}

\end{tcolorbox}

\begin{tcolorbox}[
breakable,
colback=customwhite,
colframe=customyellow,
title=Planning Error,
fontupper=\fontfamily{pcr}\selectfont\scriptsize
]

\textbf{Step 1}

\textbf{Summary:}  
Compute the sum of integers from \verb|validated_integers.txt|.

\textbf{Raw Trace:}
\begin{verbatim}
awk '{s+=$0} END{print s+0}' validated_integers.txt
-> 0
\end{verbatim}

\medskip
\textbf{Step 2}

\textbf{Summary:}  
Inspect \verb|validated\_integers.txt| contents.

\textbf{Raw Trace:}
\begin{verbatim}
cat -n validated_integers.txt
\end{verbatim}

\medskip
\textbf{\textcolor{red}{Step 3 (FAILURE STEP)}}

\textbf{Summary:}  
Runs an unrelated command counting files under /etc.

\textbf{Raw Trace:}
\begin{verbatim}
find /etc -maxdepth 1 -type f | wc -l
-> 54
\end{verbatim}

\medskip
\textbf{Step 4}

\textbf{Summary:}  
Reports the unrelated result as the final answer.

\textbf{Raw Trace:}
\begin{verbatim}
answer(54)
\end{verbatim}

\end{tcolorbox}




\clearpage

\begin{tcolorbox}[
breakable,
colback=customwhite,
colframe=customyellow,
title=Memory Limitation,
fontupper=\fontfamily{pcr}\selectfont\scriptsize
]

\textbf{Step 1}

\textbf{Summary:}  
On the My Orders page, the agent reads the latest cancelled order and correctly identifies the total cost as \$365.42.

\textbf{Raw Trace:}
\begin{verbatim}
[Step 1]
Page : My Orders

Agent: The latest cancelled order is order 000000170
dated 5/17/23 with total $365.42.

Action: click [45873]  # Video Games
\end{verbatim}

\medskip
\textbf{Step 2}

\textbf{Summary:}  
After navigating to the Video Games page, the agent behaves as if it does not know the cancelled order total and decides to open the account page to retrieve it again.

\textbf{Raw Trace:}
\begin{verbatim}
[Step 2]
Page : Video Games

Agent: To find the latest cancelled order, I need to open
the account page where the order history is located.

Action: click [48620]  # My Account
\end{verbatim}

\medskip
\textbf{Step 3}

\textbf{Summary:}  
The agent opens the order history again to search for the cancelled order, even though the total was already observed earlier.

\textbf{Raw Trace:}
\begin{verbatim}
[Step 3]
Page : My Account

Agent: I should first open the order history to find the
latest cancelled order.

Action: click [51825]  # My Orders
\end{verbatim}

\medskip
\textbf{\textcolor{red}{Step 4 (FAILURE STEP)}}

\textbf{Summary:}  
Instead of using the already known value (\$365.42), the agent continues navigating away from the orders page and repeatedly attempts to retrieve the same information, indicating that the earlier observation is no longer retained in its working memory.

\textbf{Raw Trace:}
\begin{verbatim}
[Step 4]
Page : My Orders

Agent: I will browse the Video Games category and then
report the total of the latest cancelled order.

Action: click [53661]  # Video Games
\end{verbatim}

\end{tcolorbox}

\subsection{Embodied}



\begin{tcolorbox}[
breakable,
colback=customwhite,
colframe=customgreen,
title=Instruction,
fontupper=\fontfamily{pcr}\selectfont\scriptsize
]

\textbf{Step 1}

\textbf{Summary:}  
Initialize by homing the arm.

\textbf{Raw Trace:}
\begin{verbatim}
home({})
\end{verbatim}

\medskip
\textbf{\textcolor{red}{Step 2 (FAILURE STEP)}}

\textbf{Summary:}  
Moves to a fixed pose assuming the target is the red cube without verifying the object color.

\textbf{Raw Trace:}
\begin{verbatim}
move_to_pose({'arm': 'right',
'pose': [0.2, 0.06, 1.0, 0, -0.818, 0.574, 0]})
\end{verbatim}

\medskip
\textbf{Step 3}

\textbf{Summary:}  
Attempts to grasp at the assumed target location.

\textbf{Raw Trace:}
\begin{verbatim}
envelop_grasp({'arm': 'right'})
\end{verbatim}

\medskip
\textbf{Step 4}

\textbf{Summary:}  
Lifts the object after the grasp without confirming object identity.

\textbf{Raw Trace:}
\begin{verbatim}
move_to_pose({'arm': 'right',
'pose': [0.2, 0.06, 1.1, 0, -0.818, 0.574, 0]})
\end{verbatim}

\end{tcolorbox}




\begin{tcolorbox}[
breakable,
colback=customwhite,
colframe=customgreen,
title=False Assumption,
fontupper=\fontfamily{pcr}\selectfont\scriptsize
]

\textbf{Step 1}

\textbf{Summary:}  
Return the arm to the home position.

\textbf{Raw Trace:}
\begin{verbatim}
home({})
\end{verbatim}

\medskip
\textbf{\textcolor{red}{Step 2 (FAILURE STEP)}}

\textbf{Summary:}  
Moves to a hardcoded pose assuming it corresponds to the blue cube without verifying the object.

\textbf{Raw Trace:}
\begin{verbatim}
move_to_pose({'arm': 'right',
'pose': [0.3, 0.26, 1.0, 0, -0.818, 0.574, 0]})
\end{verbatim}

\medskip
\textbf{Step 3}

\textbf{Summary:}  
Attempts to grasp whatever object is at the assumed cube location.

\textbf{Raw Trace:}
\begin{verbatim}
envelop_grasp({'arm': 'right'})
\end{verbatim}

\medskip
\textbf{Step 4}

\textbf{Summary:}  
Lifts the grasped object after the unverified pick.

\textbf{Raw Trace:}
\begin{verbatim}
move_to_pose({'arm': 'right',
'pose': [0.3, 0.26, 1.1, 0, -0.818, 0.574, 0]})
\end{verbatim}

\end{tcolorbox}

\clearpage

\begin{tcolorbox}[
breakable,
colback=customwhite,
colframe=customgreen,
title=Planning Error,
fontupper=\fontfamily{pcr}\selectfont\scriptsize
]

\textbf{Step 1}

\textbf{Summary:}  
Moves the right arm toward the first cube location.

\textbf{Raw Trace:}
\begin{verbatim}
move_to_pose({'arm': 'right',
'pose': [0.2, 0.06, 1.0, 0, -0.818, 0.574, 0]})
\end{verbatim}

\medskip
\textbf{Step 2}

\textbf{Summary:}  
Grasps the first cube with the right arm.

\textbf{Raw Trace:}
\begin{verbatim}
envelop_grasp({'arm': 'right'})
\end{verbatim}

\medskip
\textbf{\textcolor{red}{Step 3 (FAILURE STEP)}}

\textbf{Summary:}  
Attempts to move to the second cube while still holding the first cube.

\textbf{Raw Trace:}
\begin{verbatim}
move_to_pose({'arm': 'right',
'pose': [0.1, 0.16, 1.0, 0, -0.818, 0.574, 0]})
\end{verbatim}

\medskip
\textbf{Step 4}

\textbf{Summary:}  
Attempts a second grasp while the gripper is already occupied.

\textbf{Raw Trace:}
\begin{verbatim}
envelop_grasp({'arm': 'right'})
\end{verbatim}

\medskip
\textbf{Step 5}

\textbf{Summary:}  
Moves toward another cube again without releasing or placing the previous one.

\textbf{Raw Trace:}
\begin{verbatim}
move_to_pose({'arm': 'right',
'pose': [0.3, 0.26, 1.0, 0, -0.818, 0.574, 0]})
\end{verbatim}

\end{tcolorbox}







\subsection{Database}



\begin{tcolorbox}[
breakable,
colback=customwhite,
colframe=customred,
title=Instruction,
fontupper=\fontfamily{pcr}\selectfont\scriptsize
]

\textbf{Step 1}

\textbf{Summary:}  
The question requests account numbers of the oldest female clients with the lowest average salary.

\textbf{Raw Trace:}
\begin{verbatim}
Question:
Name the account numbers of female clients who are oldest
and have lowest average salary?
\end{verbatim}

\medskip
\textbf{Step 2}

\textbf{Summary:}  
The evidence clarifies schema meanings for gender and salary.

\textbf{Raw Trace:}
\begin{verbatim}
Evidence:
Female refers to 'F' in the gender;
A11 contains information about average salary.
\end{verbatim}

\medskip
\textbf{\textcolor{red}{Step 3 (FAILURE STEP)}}

\textbf{Summary:}  
The agent SQL filters only the global minimum salary district and then selects the oldest female within that subset.

\textbf{Raw Trace:}
\begin{verbatim}
Agent SQL:
SELECT A.account_id
FROM client C
JOIN disp D ON C.client_id = D.client_id
JOIN account A ON D.account_id = A.account_id
JOIN district R ON C.district_id = R.district_id
WHERE C.gender = 'F'
AND R.A11 = (SELECT MIN(A11) FROM district)
AND C.birth_date =
(
 SELECT MIN(C2.birth_date)
 FROM client C2
 JOIN district R2 ON C2.district_id = R2.district_id
 WHERE C2.gender = 'F'
 AND R2.A11 = (SELECT MIN(A11) FROM district)
);
\end{verbatim}

\medskip
\textbf{Step 4}

\textbf{Summary:}  
Gold SQL orders by \texttt{birth\_date} (oldest first) then salary and limits to one result.

\textbf{Raw Trace:}
\begin{verbatim}
Gold SQL:
SELECT T3.account_id
FROM client T1
JOIN district T2 ON T1.district_id = T2.district_id
JOIN account T3 ON T2.district_id = T3.district_id
JOIN disp T4 ON T1.client_id = T4.client_id
AND T4.account_id = T3.account_id
WHERE T1.gender = 'F'
ORDER BY T1.birth_date ASC, T2.A11 ASC
LIMIT 1;
\end{verbatim}

\medskip
\textbf{Step 5}

\textbf{Summary:}  
The system marks the agent's answer as incorrect.

\textbf{Raw Trace:}
\begin{verbatim}
Correct : False
\end{verbatim}

\end{tcolorbox}




\clearpage

\begin{tcolorbox}[
breakable,
colback=customwhite,
colframe=customred,
title=False Assumption,
fontupper=\fontfamily{pcr}\selectfont\scriptsize
]

\textbf{Step 1}

\textbf{Summary:}  
User asks for statistics about female clients in qualifying districts with salary and loan constraints.

\textbf{Raw Trace:}
\begin{verbatim}
Question:
What are the overall statistics for female clients in districts
where the average salary is between 6,000 and 10,000,
ranking in the top 3 for salary within their region,
having at least 5 female clients, and with active loan accounts?
\end{verbatim}

\medskip
\textbf{Step 2}

\textbf{Summary:}  
Evidence clarifies schema meanings for salary, gender, and loan status.

\textbf{Raw Trace:}
\begin{verbatim}
Evidence:
Average salary refers to A11
Female refers to gender = 'F'
Loan status:
'A' = active
'B' = completed
'C' = defaulted
\end{verbatim}

\medskip
\textbf{\textcolor{red}{Step 3 (FAILURE STEP)}}

\textbf{Summary:}  
The agent constructs SQL that incorrectly joins account on \texttt{district\_id} and computes unrelated birth-date statistics.

\textbf{Raw Trace:}
\begin{verbatim}
Agent SQL:
SELECT COUNT(*) AS female_count,
MIN(birth_date) AS oldest_birth_date,
MAX(birth_date) AS youngest_birth_date
FROM client
WHERE gender = 'F'
AND district_id IN ( ... )
\end{verbatim}

\medskip
\textbf{Step 4}

\textbf{Summary:}  
Gold SQL correctly computes district-level statistics using CTEs and joins.

\textbf{Raw Trace:}
\begin{verbatim}
Gold SQL:
WITH DistrictStats AS (...)
SELECT COUNT(DISTINCT ds.district_id),
SUM(ds.female_clients),
AVG(ds.avg_salary),
...
FROM DistrictStats ds
JOIN AccountActivity aa
ON ds.district_id = aa.district_id;
\end{verbatim}

\medskip
\textbf{Step 5}

\textbf{Summary:}  
The system indicates the answer is incorrect.

\textbf{Raw Trace:}
\begin{verbatim}
Correct : False
\end{verbatim}

\end{tcolorbox}

\begin{tcolorbox}[
breakable,
colback=customwhite,
colframe=customred,
title=Planning Error,
fontupper=\fontfamily{pcr}\selectfont\scriptsize
]

\textbf{Step 1}

\textbf{Summary:}  
User asks for multiple statistics about high schools in Amador County.

\textbf{Raw Trace:}
\begin{verbatim}
Question:
What are the key statistics for high schools in Amador County,
including enrollment, charter breakdown, SAT metrics,
largest school, and high-poverty counts?
\end{verbatim}

\medskip
\textbf{Step 2}

\textbf{Summary:}  
Definitions clarify high school grades, poverty threshold, and charter flag.

\textbf{Raw Trace:}
\begin{verbatim}
Evidence:
High schools: Low Grade = '9' AND High Grade = '12'
High poverty: Percent Eligible FRPM > 0.75
Charter schools: Charter School = 1
\end{verbatim}

\medskip
\textbf{\textcolor{red}{Step 3 (FAILURE STEP)}}

\textbf{Summary:}  
Agent produces only a single query counting high-poverty schools.

\textbf{Raw Trace:}
\begin{verbatim}
Agent SQL:
SELECT COUNT(*)
FROM frpm
WHERE County Name = 'Amador'
AND Low Grade = '9'
AND High Grade = '12'
AND Percent Eligible FRPM > 0.75;
\end{verbatim}

\medskip
\textbf{Step 4}

\textbf{Summary:}  
Gold SQL computes all required metrics using multiple CTEs.

\textbf{Raw Trace:}
\begin{verbatim}
Gold SQL:
WITH SchoolInfo AS (...),
SATData AS (...)
SELECT COUNT(...), AVG(...), ...
FROM SchoolInfo;
\end{verbatim}

\medskip
\textbf{Step 5}

\textbf{Summary:}  
Notes indicate the model failed to decompose the multi-part request.

\textbf{Raw Trace:}
\begin{verbatim}
Notes:
Model failed to decompose the request and produced
only a single simple query.
\end{verbatim}

\end{tcolorbox}

\begin{tcolorbox}[
breakable,
colback=customwhite,
colframe=customred,
title=Planning Error,
fontupper=\fontfamily{pcr}\selectfont\scriptsize
]

\textbf{Step 1}

\textbf{Summary:}  
User asks for multiple statistics about high schools in Amador County.

\textbf{Raw Trace:}
\begin{verbatim}
Question:
What are the key statistics for high schools in Amador County,
including enrollment, charter breakdown, SAT metrics,
largest school, and high-poverty counts?
\end{verbatim}

\medskip
\textbf{Step 2}

\textbf{Summary:}  
Definitions clarify high school grades, poverty threshold, and charter flag.

\textbf{Raw Trace:}
\begin{verbatim}
Evidence:
High schools: Low Grade = '9' AND High Grade = '12'
High poverty: Percent Eligible FRPM > 0.75
Charter schools: Charter School = 1
\end{verbatim}

\medskip
\textbf{\textcolor{red}{Step 3 (FAILURE STEP)}}

\textbf{Summary:}  
Agent produces only a single query counting high-poverty schools.

\textbf{Raw Trace:}
\begin{verbatim}
Agent SQL:
SELECT COUNT(*)
FROM frpm
WHERE County Name = 'Amador'
AND Low Grade = '9'
AND High Grade = '12'
AND Percent Eligible FRPM > 0.75;
\end{verbatim}

\medskip
\textbf{Step 4}

\textbf{Summary:}  
Gold SQL computes all required metrics using multiple CTEs.

\textbf{Raw Trace:}
\begin{verbatim}
Gold SQL:
WITH SchoolInfo AS (...),
SATData AS (...)
SELECT COUNT(...), AVG(...), ...
FROM SchoolInfo;
\end{verbatim}

\medskip
\textbf{Step 5}

\textbf{Summary:}  
Notes indicate the model failed to decompose the multi-part request.

\textbf{Raw Trace:}
\begin{verbatim}
Notes:
Model failed to decompose the request and produced
only a single simple query.
\end{verbatim}

\end{tcolorbox}







\newtcolorbox{TraceLog}[3]{
  colback=gray!5,       
  colframe=gray!60,     
  fonttitle=\bfseries,  
  title={#1 $\mid$ #2 \hfill Refs:~\cite{#3}}, 
  sharp corners,        
  boxrule=0.5pt,       
  left=2pt, right=2pt, top=2pt, bottom=2pt, 
  breakable,            
  before upper={\lstset{
    basicstyle=\ttfamily\scriptsize, 
    breaklines=true,                 
    columns=fullflexible,
    keepspaces=true,
    frame=none,                      
    aboveskip=0pt, belowskip=0pt
  }}
}

\subsection{How We Construct and Label Failure Mini-Trajectories}

\paragraph{Source and search procedure.}
Our examples are WebArena-style failure turning points expressed in the canonical web-agent interaction schema (goal + observation/accessibility-tree snippet + action), and organized by WebArena's five domains (Shopping, Shopping Admin/CMS, GitLab, Reddit, Map). We first seeded candidate failure patterns by reviewing public analyses of web-agent traces that report recurring breakdown modes such as ineffective UI interactions (no state change loops), input accumulation due to non-overwriting typing semantics, and hallucinated form fields.
We then \emph{instantiated} each seeded pattern into a domain-specific micro-scenario by (i) choosing a domain-consistent subgoal (e.g., sorting, filtering, form submission, route planning), (ii) writing a minimal observation snippet that contains the critical UI affordances and state signals needed to trigger the failure, and (iii) emitting a 2--4 step action sequence that reaches the failure point (the first irreversible divergence or loop onset). This yields a compact catalog of representative turning points while preserving the execution semantics and action space assumed by common WebArena baselines.

\paragraph{What counts as long-horizon.}
We define a task as \emph{long-horizon} if correct completion requires either (a) \textbf{multi-step dependency} where earlier constraints must be retained and enforced after intervening navigation/interaction, or (b) \textbf{multi-subgoal composition} (e.g., discover $\rightarrow$ verify $\rightarrow$ write/submit) that cannot be solved by a single local decision. Operationally, our appendix marks a failure as long-horizon when the erroneous action depends on (i) an earlier constraint or intermediate result that is no longer in the immediate observation window, or (ii) a missing prerequisite subgoal that should have been executed earlier (e.g., sorting before selecting the ``top'' comment).

\paragraph{Failure type assignment.}
We label each mini-trajectory with one of seven failure types using a lightweight decision rule that prioritizes \emph{proximal cause} at the turning point: (1) \textbf{Environment disturbance} when the UI/state changes render the agent's action ineffective or stale (e.g., modal overlays, session expiry, non-responsive dropdowns); (2) \textbf{Instruction misunderstanding} when the action violates an explicit user constraint (``do not purchase'', ``walking not driving''); (3) \textbf{Catastrophic forgetting} when an early constraint is dropped after intervening steps; (4) \textbf{False assumptions} when missing information is fabricated (e.g., user identity fields, absent emails); (5) \textbf{Planning errors} when prerequisite subgoals are skipped or ordered incorrectly; (6) \textbf{History error accumulation} when small state-tracking errors compound (e.g., append-typing, toggle oscillation); and (7) \textbf{Memory limitations} when the agent cannot retain or aggregate required multi-item evidence across steps. These categories are consistent with diagnostic perspectives that seek to attribute failures to specific breakdown loci in interactive web settings.

\paragraph{Why we only show 2 - 4 steps.}
Although the underlying tasks are long-horizon, the \emph{failure turning point} is typically localized: a brief window where the agent first enters an unrecoverable branch (constraint violation) or an absorbing loop (no-state-change / oscillation), which makes a 2 - 4 step slice an adequate witness of the long-horizon breakdown.
\begin{lstlisting}[
  basicstyle=\ttfamily\footnotesize,
  frame=single,
  breaklines=true,
  breakatwhitespace=true,
  postbreak=\mbox{\textcolor{gray}{$\hookrightarrow$}\space}
]
if (action has no effect OR UI state changed/stale element/overlay/session): ENV_DIST
else if (action violates explicit constraint OR wrong modality/target): INSTR_MIS
else if (fabricates missing fields OR claims unseen facts): FALSE_ASSUME
else if (drops earlier constraint / missing prerequisite / compounding drift): FORGET/PLAN/HIST/MEM by nearest cause
\end{lstlisting}

\clearpage
\section{Appendix: LLM-as-a-Judge Result} \label{app:4}

This appendix presents the full quantitative results of our trajectory-grounded LLM-as-a-Judge
failure attribution pipeline across all domains and models.
Across 3,132 trajectories (1,995 failures, 1,137 successes), the overall task success rate is
\textbf{36.3\%}, with substantial variation across domains and models.
By domain, success rates are: Web 24.1\% (185/767), OS 40.3\% (864/2,146),
Database 36.9\% (31/84), and Embodied 42.2\% (57/135).
By model, GPT-5-mini achieves 33.4\% (573/1,718) and Claude-4-Sonnet achieves 39.9\% (564/1,414).
The two models exhibit markedly different failure profiles: GPT failures are dominated by
Planning Errors (64.9\%) and Memory Limitation (18.3\%), while Claude failures show a higher
proportion of Environment failures (32.5\%) and Instruction failures (16.5\%), with Planning
Errors still the most common (46.5\%).

We validate the judge against expert labels on a pilot set of 40 trajectories.
On this set, inter-annotator agreement is \(\kappa{=}0.61\) and human--judge agreement is \(\kappa{=}0.84\).

\begin{figure}[h]
  \centering
  \includegraphics[width=\textwidth]{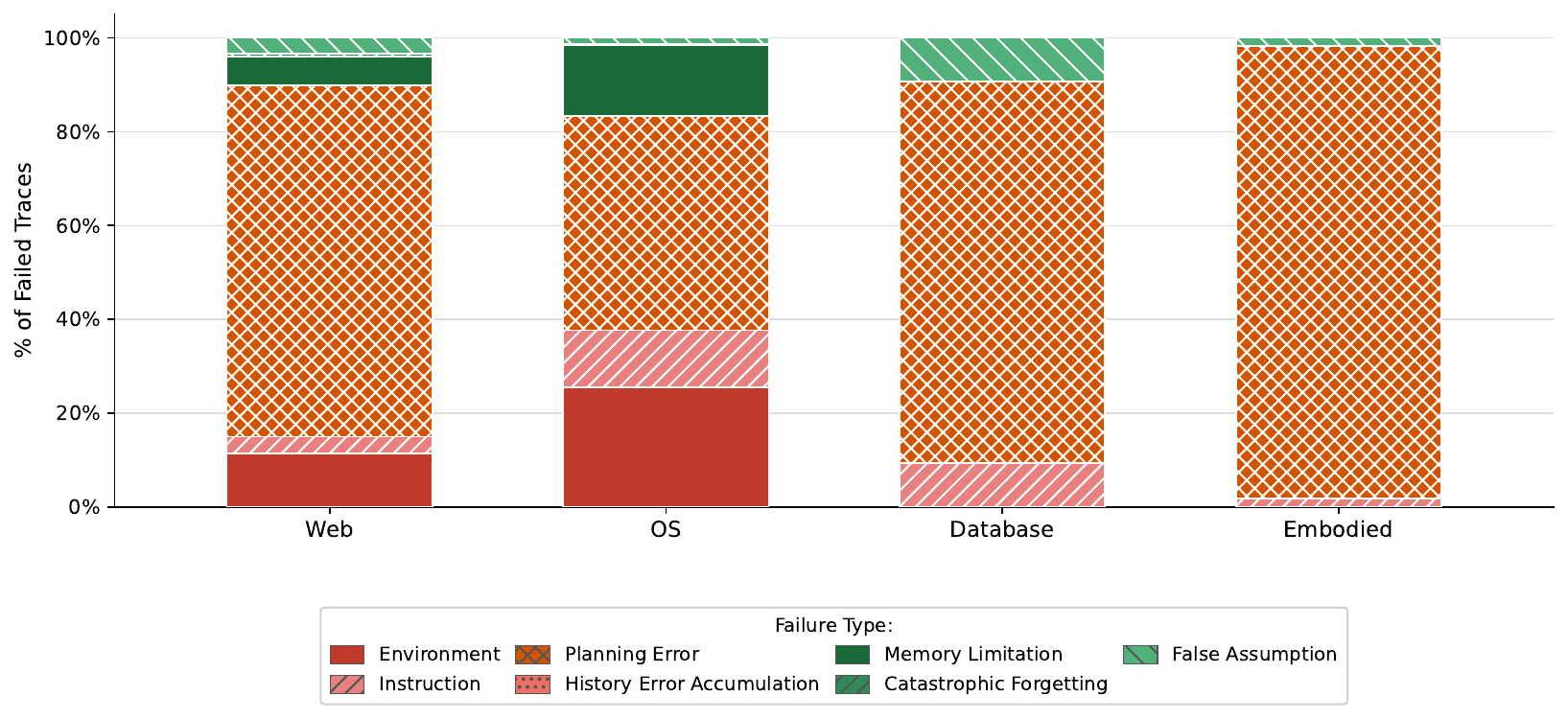}
  \caption{\textit{Failure mode distribution across the four task domains
  (SR = task success rate; $n$ = number of failed traces).}
  Embodied (SR 42.2\%) and Database (SR 36.9\%) are almost entirely dominated by
  Planning Error (94.9\% and 79.3\%, respectively), indicating that structured,
  well-defined action spaces surface planning as the primary bottleneck.
  Web (SR 24.1\%) follows a similar pattern (Planning Error 74.9\%) but additionally
  exhibits Environment (11.3\%) and Memory Limitation (6.2\%) failures, reflecting
  dynamic page-state variability and multi-tab navigation demands.
  OS (SR 40.3\%) has by far the most diverse failure profile: Planning Error (36.7\%),
  Instruction (25.9\%), Environment (17.3\%), and Memory Limitation (15.1\%) all
  contribute substantially, consistent with the open-ended, long-horizon nature of
  shell tasks. History Error Accumulation appears exclusively in OS (0.1\%).}
  \label{fig:appendix_by_domain}
\end{figure}

\begin{figure}[!tb]
  \centering
  \includegraphics[width=0.7\textwidth]{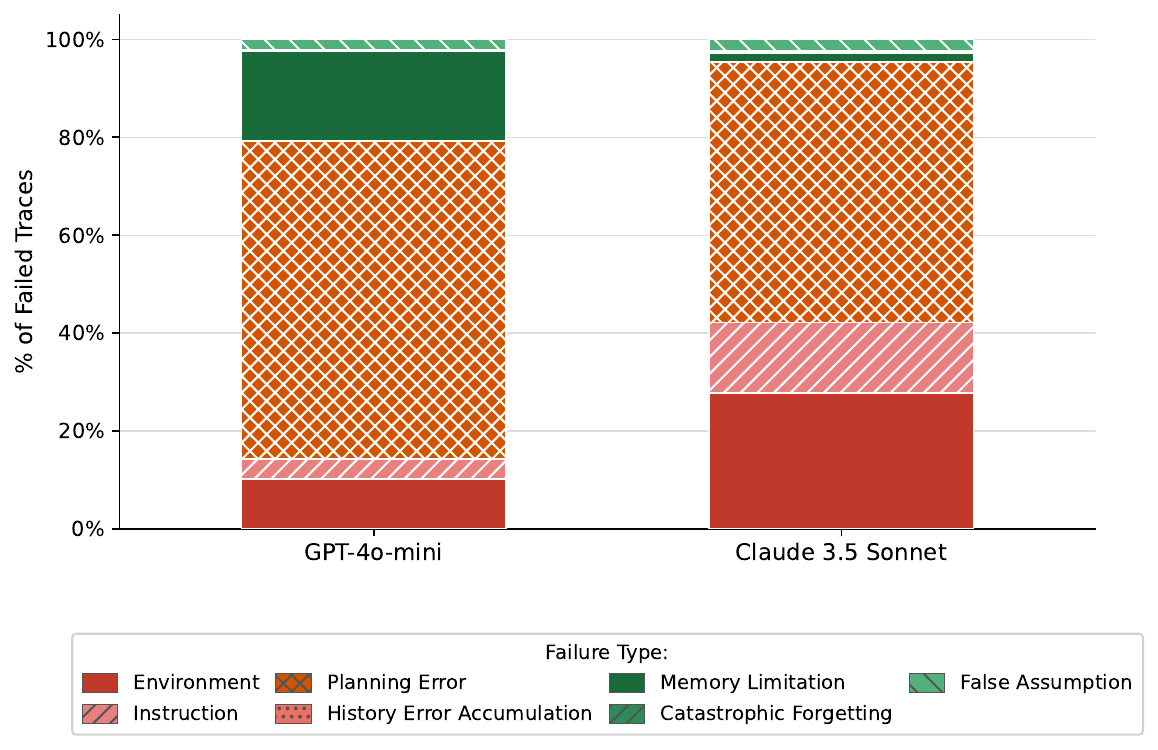}
  \caption{\textit{Failure mode distribution for GPT-4o-mini vs.\ Claude~3.5 Sonnet
  aggregated across all domains.}
  The two models exhibit qualitatively different failure regimes.
  GPT-4o-mini (1,145 failed traces, success rate 33.4\%) is dominated by Planning Error (64.9\%) and Memory
  Limitation (18.3\%), reflecting difficulty maintaining coherent long-horizon plans
  and retaining intermediate state across extended rollouts.
  Claude~3.5 Sonnet (850 failed traces, success rate 39.9\%) shows markedly higher Environment (32.5\%) and
  Instruction (16.5\%) failures while Memory Limitation is near-zero (2.2\%),
  suggesting stronger context retention but greater sensitivity to environment-state
  changes and task-specification ambiguity.
  Design-level risks (Memory Limitation, Catastrophic Forgetting, False Assumption)
  account for 20.8\% of GPT failures but only 6.6\% of Claude failures.}
  \label{fig:appendix_by_model}
\end{figure}

\begin{figure}[!tb]
  \centering
  \includegraphics[width=\textwidth]{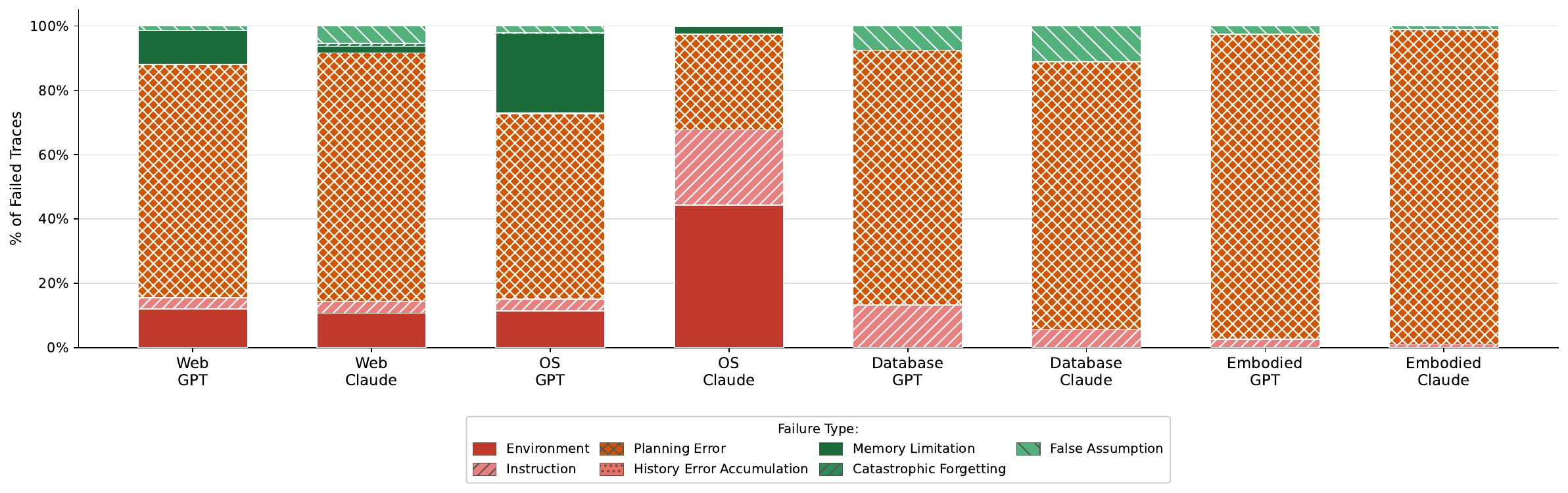}
  \caption{\textit{Failure mode distribution broken down by domain and model.}
  Database and Embodied only have GPT trajectories in the current dataset.
  \textbf{Web:} Both models are planning-error dominated ($>$72\%), but GPT accumulates
  more Memory Limitation (10.6\% vs.\ 2.0\%) while Claude shows higher False Assumption
  (5.4\% vs.\ 1.4\%), pointing to distinct secondary failure mechanisms.
  \textbf{OS:} The model divergence is most pronounced here. GPT failures concentrate in
  Planning Error (57.7\%) and Memory Limitation (24.6\%), consistent with context-length
  degradation over long command chains. Claude failures are instead driven by
  Environment (44.3\%) and Instruction (23.4\%), indicating sensitivity to
  shell-environment feedback and command-specification ambiguity.
  \textbf{Database:} GPT shows Planning Error (79.3\%) and False Assumption (7.5\%)
  as the main failure modes, with no Memory Limitation.
  \textbf{Embodied:} Nearly all failures are Planning Errors (94.9\%),
  reflecting the precise sequential action ordering demands of embodied tasks.}
  \label{fig:appendix_by_domain_model}
\end{figure}

\clearpage
\section{Appendix: More Discussion}

This work aims to advance nuanced discussions of systematic long-horizon analysis and to provide methodological guidance for scalable, reproducible, and cross-domain research on long-horizon agentic AI. 
Below, we discuss the implications of findings in Section~\ref{sec:5} and outline directions for future research. 

\textbf{Empirical Support for \texttt{HORIZON}.}
First, we observe a consistent pattern of performance collapse beyond early horizon extension levels across all four domains, providing mechanistic evidence that increasing $s$ indeed pushes agents into regimes dominated by long-horizon reasoning failure. This sharp transition indicates that our horizontal horizon construction meaningfully orders tasks by long-horizon difficulty. 
In parallel, qualitative analysis of failed trajectories shows that failures can be systematically categorized using the seven-category taxonomy defined in \texttt{HORIZON} (Table~\ref{tab:failure_tax}), providing phenomenological structure to observed breakdowns. Notably, all annotated failures observed across domains in our experiments fall within this taxonomy, despite substantial differences in task modality and environment dynamics. This suggests that \texttt{HORIZON} provides a valid and generalizable framework for studying long-horizon tasks across domains.

\textbf{Long-Horizon Tasks Lack a Canonical Definition.}
A central challenge highlighted by our study is that long-horizon tasks remain poorly defined in prior literature. Existing benchmarks operationalize ``horizon'' using incompatible proxies, such as interaction length, number of tool calls, reasoning steps, or episode duration, often without explicit justification. As a result, what constitutes a long-horizon task in one domain may be trivial, incomparable, or ill-defined in another. Our findings reinforce that horizon cannot be normalized by step count alone and must instead be grounded in task structure. This motivates the need for principled, task-centric definitions, as adopted in \texttt{HORIZON}, to enable meaningful cross-domain comparison.

\textbf{Long-Horizon Failure Reflects Fundamental Generalization Limits.}
Long-horizon failures should not be viewed merely as extensions of short-horizon planning errors. Instead, they reflect a distinct form of generalization under distribution shift, where agents must operate under compounding changes in state, action consequences, and goal dependencies. As horizons increase, agents encounter states and decision contexts that diverge increasingly from those seen earlier in the trajectory. We suggest that breakdowns in long-horizon settings arise not only from insufficient planning depth, but from systematic limits in maintaining and updating beliefs over extended interactions. This reframes long-horizon reasoning as a core generalization challenge rather than a purely algorithmic planning problem.

\textbf{Why Scaling Alone Is Unlikely to Solve Long-Horizon Failures.}
Although frontier models outperform smaller systems at moderate horizons, our results show that performance gaps often collapse once agents enter the long-horizon failure regime. This convergence suggests diminishing returns from additional model capacity alone. Importantly, different failure types demand qualitatively different interventions: environment disturbances require monitoring and recovery mechanisms, catastrophic forgetting calls for improved memory and constraint tracking, while planning errors may benefit from better sub-goal decomposition. Therefore, one-size-fits-all solution, such as increasing model size or training on longer traces, is unlikely to address the root causes of long-horizon failure. Systematic diagnosis is a prerequisite for targeted improvement.

\textbf{Breaking Points as Transition Regions, Not Thresholds.}
Our empirical findings support viewing breaking points as transition regions rather than precise thresholds. Across domains, performance collapse occurs abruptly but at different horizons, and often spans a narrow range of extension levels rather than a single step. This reconciles the intuitive appeal of breaking points with our theoretical argument against universal definitions. Operationally, it suggests that evaluation should focus on identifying regions where failure dynamics qualitatively change, rather than attempting to pinpoint exact breaking steps. Such region-based analysis is more robust to domain variation and model heterogeneity.

\textbf{Implications for Benchmark Design and Agent Evaluation.}
Taken together, our results call for a shift in how long-horizon agents are evaluated. Reporting single-point accuracy obscures where and how failures emerge as horizons grow. Horizon-aware evaluation—reporting performance curves, failure transitions, and attributed failure types—provides a richer and more actionable picture of agent behavior. By decoupling horizon measurement from failure attribution, \texttt{HORIZON} offers a flexible framework for studying long-horizon reasoning without assuming a universal breaking point. We hope this perspective encourages the community to design benchmarks and diagnostics that better reflect the realities of long-horizon agent deployment.

\end{document}

%% file: fig_agent.tex
\begin{figure*}[!tb]
  \includegraphics[width=\textwidth]{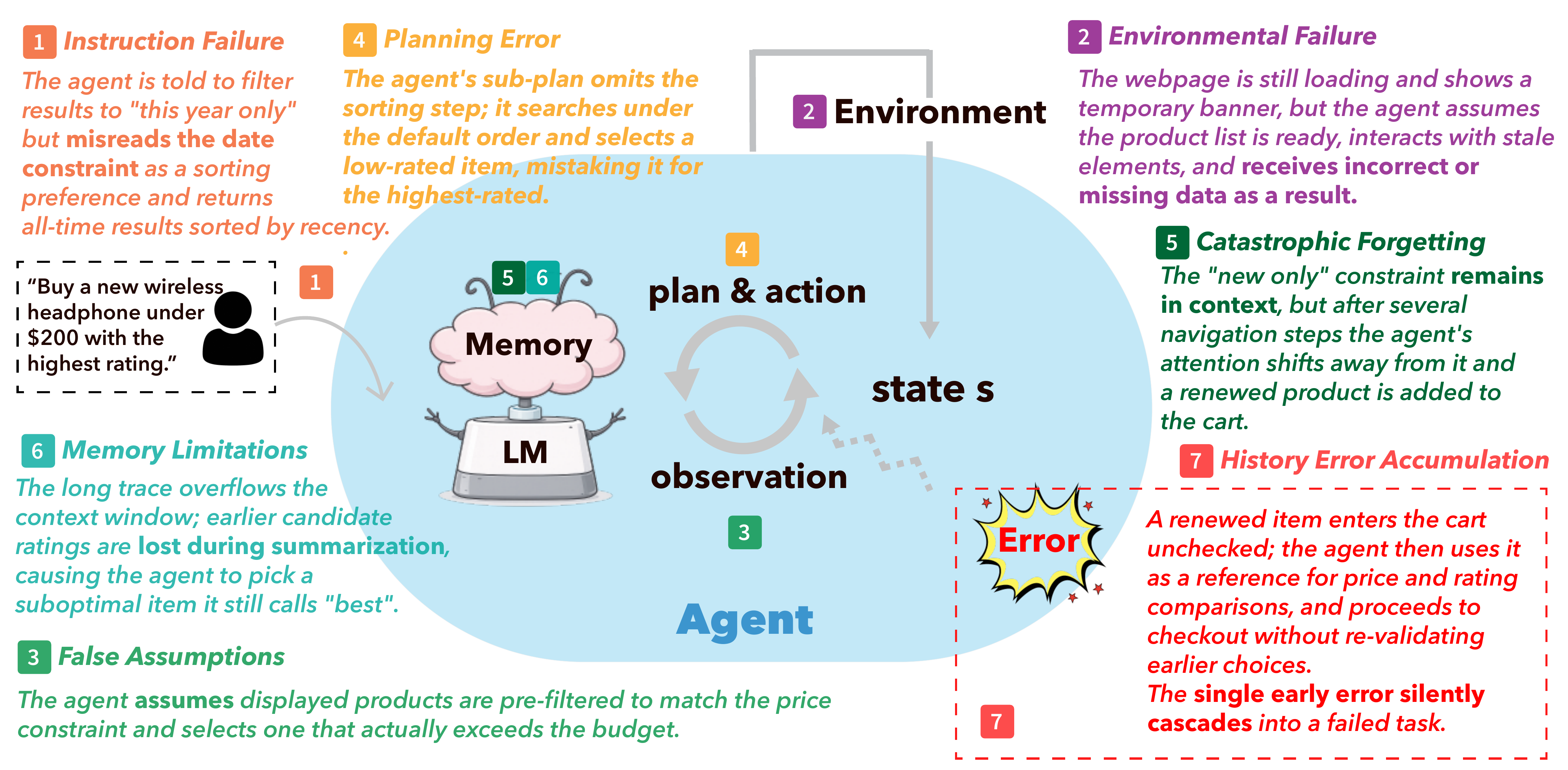}
  \centering
\caption{\textit{Illustration of general agent execution and failure propagation.}
Given an instruction \instbox, the agent iterates a standard loop:
it observes the environment \envbox to obtain observations \obsbox,
plans and selects an action \planbox drawing on memory-backed context \membox,
executes that action to change the environment,
and then updates its internal state \forgetbox through memory \membox.
Failures can originate at any stage and compound across cycles \histbox.
The seven legend boxes map directly to the failure categories in Section~\ref{subsec:failure_taxonomy}.}

  \label{fig:agent_failure_framework}
\end{figure*}

%% file: fig_workflow.tex
\begin{figure*}[!tb]
  \includegraphics[width=\textwidth]{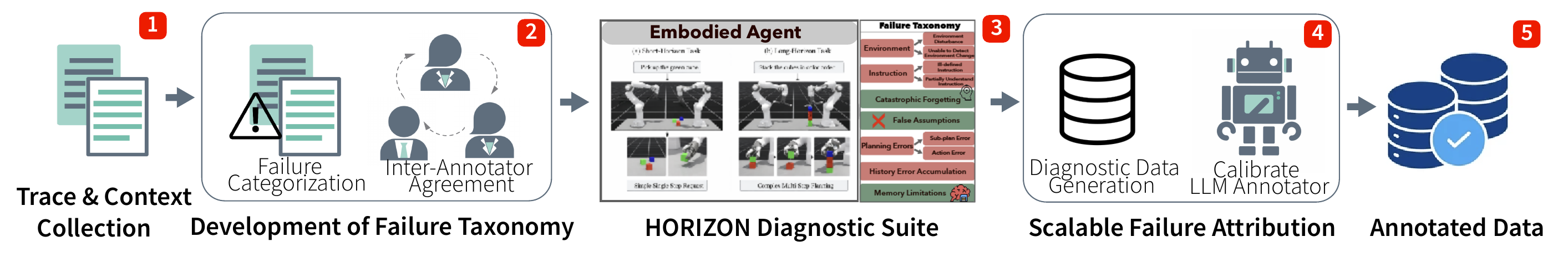}
  \centering
   \caption{\textit{The \texttt{HORIZON} diagnostic pipeline for scalable long-horizon failure analysis.} The pipeline consists of trace and context collection, taxonomy development with inter-annotator agreement, construction of the \texttt{HORIZON} diagnostic suite (unified horizon measurement and failure attribution), calibration of an LLM-based annotator, and large-scale failure annotation. This figure highlights our systematic approach to approximating the breaking-point problem in long-horizon agents, enabling scalable and reproducible analysis for future cross-domain research. More details are provided in Section~\ref{sec:4}.
}
  \label{fig:workflow}
\end{figure*}

%% file: fig_horizon.tex
\begin{figure*}[!tb]
  \includegraphics[width=\textwidth]{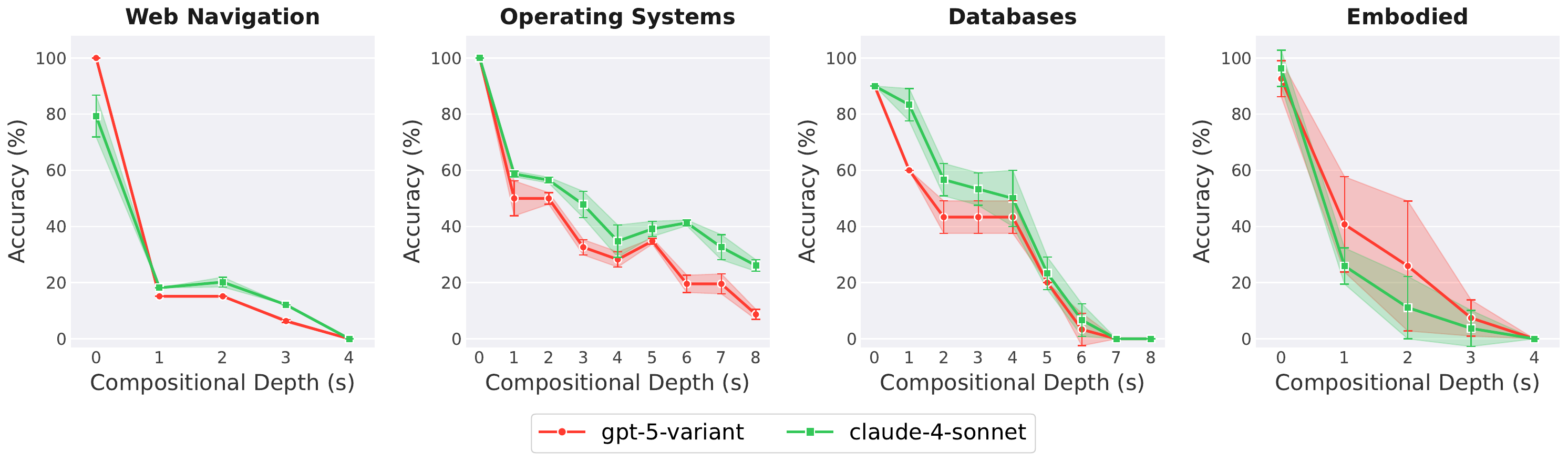}
  \centering
   \caption{\textit{Current model performance as a function of horizon extension.} Plots show success rate (accuracy) versus the compositional depth $s$, where $s$ denotes the \texttt{HORIZON}-defined extension level corresponding to the number of high-level subtasks composed within a task. Accuracy is computed over task sets with the same $s$. Each point reports the mean over three independent runs with identical task sets; variability across runs is shown as $\pm$ one standard deviation (error bars). We evaluate GPT-5-mini and Claude-4-Sonnet across all domains. Across models and domains, performance consistently decreases and exhibits a sharp drop beyond a certain value of $s$, empirically supporting the validity of our horizontal horizon definition. More details are provided in Section~\ref{sec:5}.}
  \label{fig:horizon}
\end{figure*}

%% file: fig_teaser.tex
\begin{figure*}[!tb]
  \includegraphics[width=\textwidth]{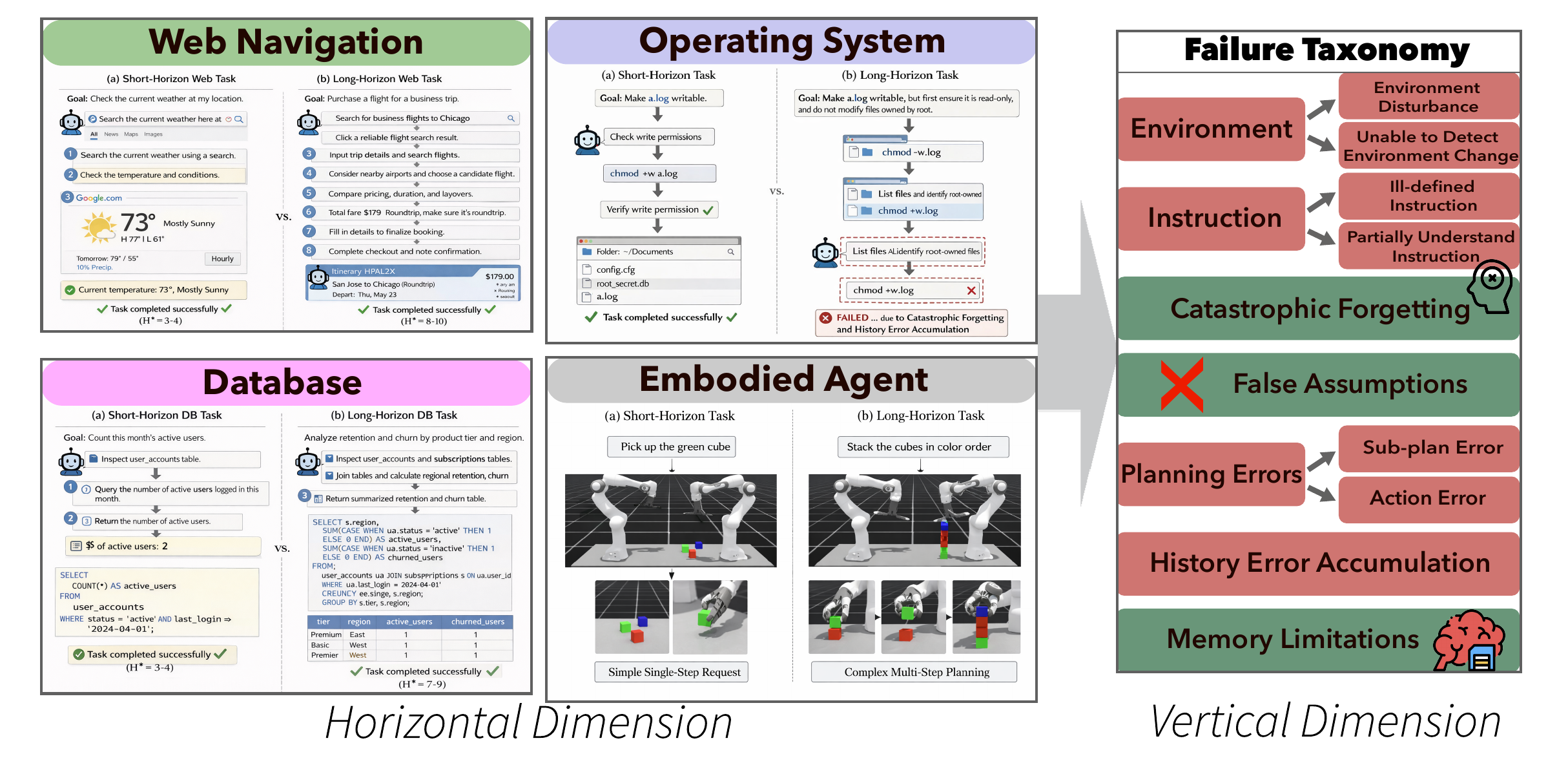}
  \centering
   \caption{\textit{\texttt{HORIZON} overview with two orthogonal dimensions.} \textit{Left} (Horizontal / Horizon): four domain examples contrasting short- vs. long-horizon tasks under our task-structure definition, where intrinsic horizon $H^*$ increases with extension level $s$ (and may also increase compositional depth $C$). From top-left to bottom-right: Web illustrates theoretical $H^*$ growth (e.g., “purchase a flight,” $H^*=8$, vs. “check the weather,” $H^*=3$); OS shows depth extension by inserting non-skippable intermediate states (e.g., enforce and verify read-only before making \texttt{a.log} writable); Database and Embodied show breadth/length extension by composing multiple short-horizon subtasks into one workflow, increasing number of subtasks and coordination steps. \textit{Right} (Failure Attribution): 7-category failure taxonomy. 
    Colors indicate a hierarchical risk perspective inspired by FMEA: blue corresponds to design-level risks (DFMEA), originating from limitations in agent architecture, while orange corresponds to process-level risks (PFMEA), emerging during sequential execution.
    More details are provided in Section~\ref{sec:horizon}. 
}
  \label{fig:teaser}
\end{figure*}